\documentclass[times,twocolumn,final,authoryear]{elsarticle}

\RequirePackage{geometry}
\usepackage{framed,multirow}
\usepackage[table,xcdraw]{xcolor}

\usepackage{url}
\usepackage{xcolor}
\definecolor{newcolor}{rgb}{.8,.349,.1}

\usepackage{placeins}
\usepackage{amsfonts}
\usepackage{amssymb}
\usepackage{amsmath}
\usepackage{rotating}
\usepackage{multirow}
\usepackage[pdftex]{hyperref}
\hypersetup{pdfauthor={Name}}

\journal{Computer Vision and Image Understanding}

\begin{document}

\begin{frontmatter}

    \title{Iterative saliency enhancement over superpixel similarity}

    \author[inst1]{Leonardo~de~Melo~Joao\corref{cor1}}
    \author[inst1]{Alexandre~Xavier~Falc{\~{a}}o}
    \address{Institute of Computer Science, University of Campinas, Campinas, 13083-872, São Paulo, Brazil}
    \cortext[cor1]{leomelo168@gmail.com}

    \begin{abstract}
Saliency Object Detection (SOD) has several applications in image analysis. The methods have evolved from image-intrinsic to object-inspired (deep-learning-based) models. When a model fail, however, there is no alternative to enhance its saliency map. We fill this gap by introducing a hybrid approach, named \textit{Iterative Saliency Enhancement over Superpixel Similarity} (ISESS), that iteratively generates enhanced saliency maps by executing two operations alternately: object-based superpixel segmentation and superpixel-based saliency estimation -- cycling operations never exploited. ISESS estimates seeds for superpixel delineation from a given saliency map and defines superpixel queries in the foreground and background. A new saliency map results from color similarities between queries and superpixels at each iteration. The process repeats and, after a given number of iterations, the generated saliency maps are combined into one by cellular automata. Finally, the resulting map is merged with the initial one by the maximum bewteen their average values per superpixel. We demonstrate that our hybrid model can consistently outperform three state-of-the-art deep-learning-based methods on five image datasets.

%Saliency Object Detection (SOD) has several applications in image analysis. Deep-learning-based methods are among the most effective, but they may miss foreground parts. To circumvent the problem, we propose a hybrid model that combines a deep-object-inspired and an image-intrinsic model. We introduce a novel image-intrinsic model, named \textit{Iterative Saliency Enhancement over Superpixel Similarity} (ISESS), that executes two operations alternately for saliency completion: object-based superpixel segmentation and superpixel-based saliency estimation. ISESS uses pre-computed saliency maps to estimate seeds for superpixel delineation and define superpixel queries in the foreground and background. A new saliency map results from color similarities between queries and superpixels at each iteration. After a given number of iterations, the saliency maps are combined into one by cellular automata. Finally, the image-intrinsic and initial maps are merged using their average superpixel values. We demonstrate that the hybrid model can consistently improve three deep-learning-based methods on five image datasets.
    \end{abstract}
    
    %Research highlights
%    \begin{highlights}
%    \item Deep-saliency maps to define queries for image-intrinsic saliency methods.
%    \item Image-intrinsic information to improve deep-saliency models.
%    \item Hybrid deep-object-inspired and image-intrinsic model for better saliency estimation.
%    \item Saliency enhancement by alternate executions of saliency and superpixel methods.  
%    \end{highlights}


\end{frontmatter}

%\maketitle

%%%% ADD BOOTSTRAP RELATED WORK, SELL THE IDEA THE ISESS IS THE ONLY METHOD THAT ENHANCES SALIENCY OF A SINGLE INPUT SALIENCY METHOD 
\section{Introduction}\label{sec:intro}
    Saliency Object Detection (SOD) aims to identify the most visually relevant regions within an image. SOD methods have been used in an extensive range of tasks, such as image segmentation \citep{iqbal2020saliency}, compression \citep{wang2021focus}, quality assessment \citep{liu2012towards}, and content-based image retrieval \citep{al2021saliency}.

  Traditional SOD methods can combine heuristics to model object-inspired (top-down) and image-intrinsic (bottom-up) information. Object-inspired models expect that salient objects satisfy specific priors based on domain knowledge -- e.g., salient objects in natural images are expected to be centered \citep{cheng2014global}, focused \citep{jiang2013salient}, or have vivid colors \citep{peng2016salient}. Image-intrinsic models provide candidate background and foreground regions used as queries and expect that salient regions be more similar to the foreground than background queries~\citep{yang2013saliency, wu2018salient}. Thus, their results strongly rely on those preselected assumptions. 

  Recently, deep-learning-based SOD methods have replaced preselected assumptions with examples of salient objects from a given training set. Such object-inspired methods usually rely on the backbone of a Convolutional Neural Network (CNN) trained for image classification. However, there are network architectures that can be easily trained from scratch~\citep{qin2020u2}.
  
\begin{figure}[t!]
        \centering
        \begin{tabular}{c c c c}
                 \includegraphics[width=0.09\textwidth]{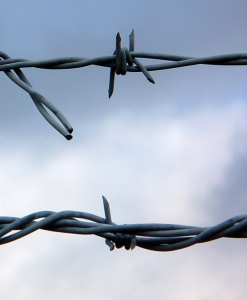} &
                 \includegraphics[width=0.09\textwidth]{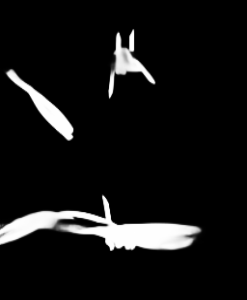} &
                 \includegraphics[width=0.09\textwidth]{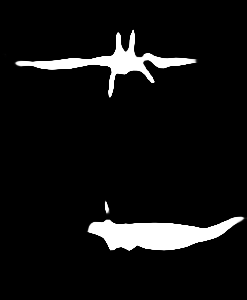} &
                 \includegraphics[width=0.09\textwidth]{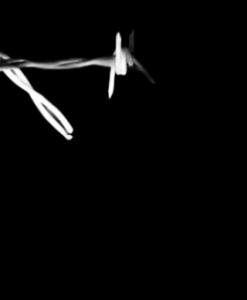}  \\
                 \includegraphics[width=0.09\textwidth]{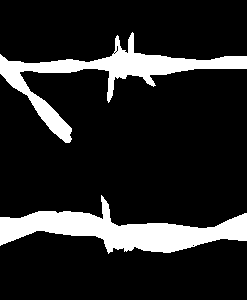} &
                 \includegraphics[width=0.09\textwidth]{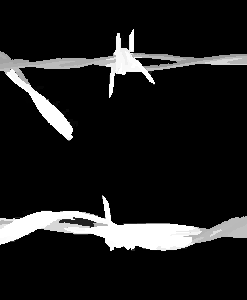} &
                 \includegraphics[width=0.09\textwidth]{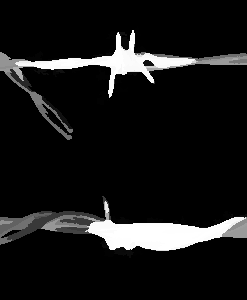} &
                 \includegraphics[width=0.09\textwidth]{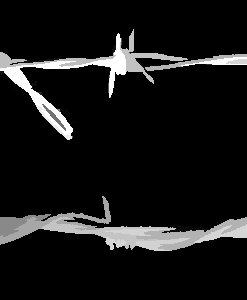}  \\
                 (a) & (b) & (c) & (d)
            \end{tabular}
        \caption{(a) Original image (top) and ground-truth segmentation (bottom). (b)-(d) BASNET \citep{qin2019basnet} MSFNet \citep{miao2021automsf}, and U²Net \citep{qin2020u2} saliency maps (top) and their enhanced maps (bottom) using the proposed approach.}
        \label{fig:intro-deep-learning-plus-sess}
\end{figure}

  Deep SOD methods are among the most effective, but they may miss foreground parts with similar colors (see Figure \ref{fig:intro-deep-learning-plus-sess}). Such incomplete saliency maps drift away from the results humans expect and there is no alternative to improve saliency maps. Since CNNs are object-inspired models that do not explore image-intrinsic information, we fill that gap by proposing a hybrid model, named  \textit{Iterative Saliency Enhancement over Superpixel Similarity} (ISESS), which explores superpixels in multiple scales and color similarity.

\begin{figure}[t!]
    \centering
        \begin{tabular}{c c c}
                 \includegraphics[width=0.12\textwidth]{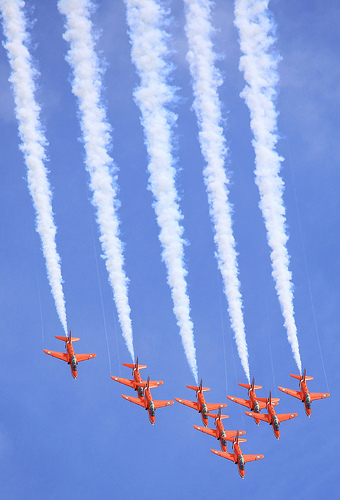} &
                 \includegraphics[width=0.12\textwidth]{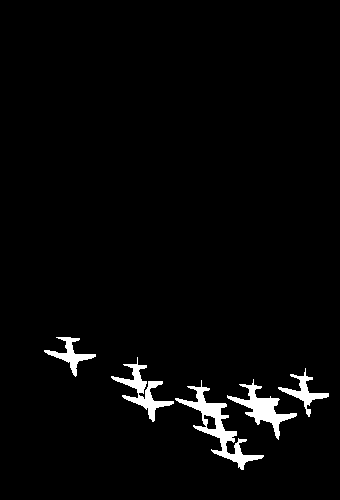} &
                 \includegraphics[width=0.12\textwidth]{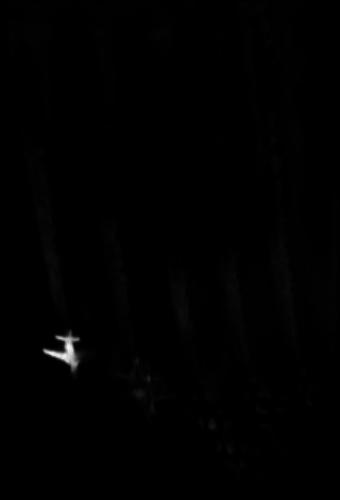} \\
                 (a) & (b) & (c)\\
                 \includegraphics[width=0.12\textwidth]{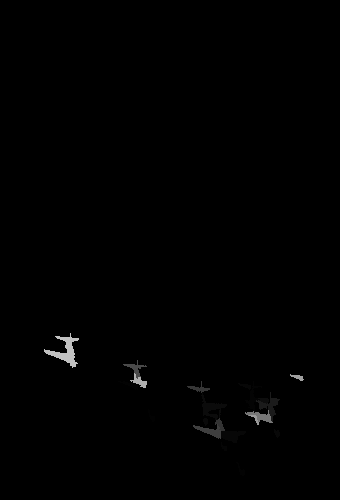} &
                 \includegraphics[width=0.12\textwidth]{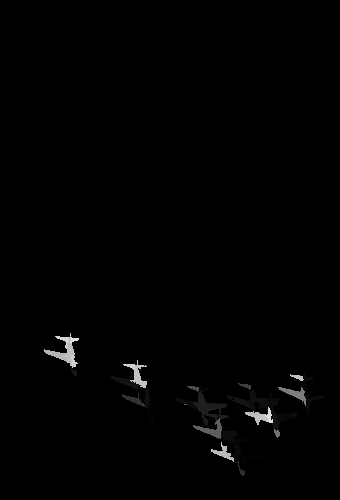} &
                 \includegraphics[width=0.12\textwidth]{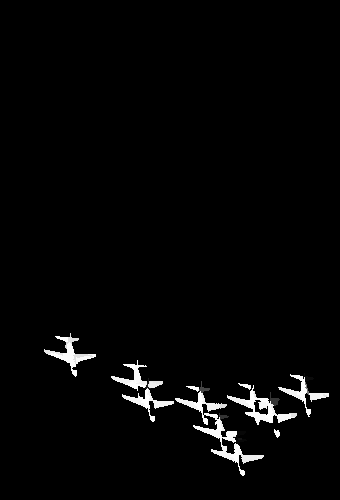} \\
                 (d) & (e) & (f)
            \end{tabular}
        \caption{ISESS saliency enhancement over multiple iterations. (a-c) original image, ground-truth segmentation and U²net saliency; (d-f) ISESS improvement over iterations 1, 3 and 10.}
        \label{fig:iteration-progress}
\end{figure}

ISESS can improve an input salience map over a few iterations of two alternate operations: object-based superpixel segmentation~\citep{belem2019oisf} and superpixel-based saliency estimation. ISESS uses an input saliency map to estimate seeds for superpixel delineation and define superpixel queries in the foreground and background. A new saliency map is obtained by measuring color similarities between queries and superpixels, such that salient superpixels are expected to have high color similarity with at least one foreground query and low color similarity with most background queries. The process repeats for a given number of iterations, with a decreasing quantity of superpixels (increasing scale) per iteration. Although the process is meant to create progressively better saliency maps (Figure~\ref{fig:iteration-progress}), we combine all output maps by cellular automata~\citep{qin2015saliency}. By using the deep model as input, ISESS removes the need for preselected assumptions, providing an image-intrinsic extension of the provided object-inspired model. Finally, the image-intrinsic and initial (object-inspired) maps are merged by the maximum between their average values per superpixel, creating a hybrid saliency model. It is worth noting that ISESS does not depend on the choice of the object-inspired method. 

    We demonstrate that our hybrid model consistently can outperform three state-of-the-art deep SOD methods (namely Basnet \citep{qin2019basnet}, U²Net \citep{qin2020u2}, and Auto-MSFNet\citep{ miao2021automsf}) using five well-known datasets. These results are confirmed by several metrics in all datasets, especially when images have multiple salient objects. Qualitative analysis also shows considerably enhanced maps whenever deep models fail to capture salient regions. 
    
    Therefore, the contributions of this paper are:
    \begin{itemize}
        \item A hybrid model for saliency object detection that does not rely on preselected assumptions while using object-inspired and multiscale image-intrinsic information.
        \item A first method to enhance object saliency maps, even when they are generated by deep-learning-based methods, with no need to combine multiple SOD methods (Section~\ref{subsec:saliency-aggregation}). 
        \item A novel strategy that alternates over time object-based superpixel segmentation and superpixel-based saliency estimation (cycling operations never exploited) for saliency estimation.
    \end{itemize}
    
    Section \ref{sec:related-works} presents related work and ISESS is detailed in Section \ref{sec:sal-enhance}. Experiments with in-depth analysis of their results are presented in Section \ref{sec:results}. We state conclusions and discuss future work in Section \ref{sec:conclusion}. 
\section{Related Work}\label{sec:related-works}

    \subsection{Traditional SOD methods}\label{subsec:traditional-methods}
    
    \subsubsection{Overview} 
    
    Traditional SOD methods often rely on hand-crafted heuristics to model a set of characteristics shared by visually salient objects. These methods explore a combination of top-down object-inspired information --- characteristics inherent to the object, not its relationship to other image regions --- or bottom-up image-intrinsic strategies that use intrinsic image features to estimate saliency based on region similarity. Common top-down approaches expect salient objects centralized \citep{cheng2014global}, composed of vivid colors \citep{peng2016salient} or focused \citep{jiang2013salient}. Early bottom-up strategies include modeling saliency according to pixel-level similarity to the mean image color \citep{achanta2009frequency} or defining a global contrast by comparing all possible pairs of image patches \citep{cheng2014global}. Later, saliency estimators started using superpixels for better region representation, consistently outperforming pixel and patch-based methods. Most superpixel-based  SOD methods use different heuristics to assemble a combination of top-down and bottom-up approaches.
    
    \subsubsection{Image-intrinsic superpixel-based approaches}
    
    Apart from superpixel similarity methods that use global contrast \citep{jiang2013salient}, bottom-up approaches (image-intrinsic) require a strategy to select superpixels that represent candidate foreground and background regions --- namely, queries. The three commonest query selection strategies are based on \textit{sparsity}, \textit{backgroundness}, or \textit{objectness}. Methods based on low-rank (LR) matrix recovery \citep{lang2011saliency, peng2016salient} use \textit{sparsity}, and assume an image can be divided into a highly redundant background-likely low-rank matrix and a salient sparse sensory matrix. Superpixel-graph-based methods \citep{zhu2014saliency, zhang2018hypergraph, yang2013saliency, wu2018salient} estimate saliency by combining superpixel-adjacency contrast with the similarity between every superpixel and probable background and foreground \textit{queries}. Background queries are usually defined at the image borders, but the methods propose strategies to mitigate error when this assumption is invalid. Such strategies include estimating four saliency maps (one for each image border) so that regions consistently highlighted are taken as salient \citep{yang2013saliency}; using only border regions that share common characteristics amongst themselves \citep{wu2018salient}; include a weight function for the query regions, so that superpixels that grow close to the image center have a lower weight than the ones connected to the image limits \citep{zhu2014saliency}. However, background-based saliency methods often highlight parts of the desired object even when using those error-mitigating strategies. To improve upon background-based saliency, one can use background saliency to define salient regions as foreground queries and then compute a foreground saliency score \citep{yang2013saliency, wu2018salient}.
    
    A major drawback in all methods above is their reliance on heuristics or domain-specific prior information. We combine saliency estimations from a given object-inspired deep SOD model and the proposed superpixel-based image-intrinsic model to avoid the heuristic-based approach. Additionally, the proposed method implements a novel enhancement loop that iteratively improves object representation by alternating executions of object-based superpixel segmentation and saliency estimation. The iterative aspect of our method explores multiple superpixel scales, which allows us to not impose spatial constraints as superpixel-graph-based methods do. We simply compute feature similarity between all superpixels and queries, leaving the pixel adjacency relationship to be handled by the different superpixel segmentations.

\subsection{Saliency map improvement by aggregation}\label{subsec:saliency-aggregation}
    Multiple SOD methods have been combined to improve incomplete saliency maps~\cite{qin2015saliency, chen2018saliency,singh2020saliency,li2018saliency}. Unsupervised methods may use min-max operations over the estimated foreground and background regions \cite{singh2020saliency}, Baysian Frameworks for iteratively enhancing consistently salient regions over time \citep{qin2015saliency}, or DS evidence theory to learn weights for the multiple saliency maps in a fusion framework \citep{chen2018saliency}. Supervised methods often learn regressors that combine multiple saliency maps into a single saliency score using bootstrap learning \cite{tong2015salient, li2018saliency}.
     
     However, such aggregation strategies create a significant processing overhead due to the execution of multiple SOD methods, and their high precision depends on an agreement among most saliency maps with respect to the salient object. To our knowledge, our approach is the first that improves incomplete saliency maps without requiring aggregation of multiple SOD methods.
    
\subsection{Deep-learning methods}
    \subsubsection{Multi-layer perceptron (MLP)-based approaches} 
    
    The usage of deep neural networks for saliency detection has been extensive in the past few years. As presented in a recent survey \citep{wang2021salient}, earlier attempts used MLP-based approaches \citep{zhang2016unconstrained, he2015supercnn, liu2016dhsnet}, adapting networks trained for image classification by appending the feature-extraction layers to a pixel-patch classifier. Despite the improvement of these models over heuristic-based traditional methods, they were incapable of providing consistent high spatial accuracy, primarily due to relying on local patch information.
    
    \subsubsection{Fully-Convolutional Neural-Networks (FCNN)}
    
    FCNN-based methods improved object detection and delineation, becoming the most common network class for visual saliency estimation. Most FCNN models use the pretrained backbone of a CNN for image classification (\textit{e.g.} DenseNet \citep{huang2017densely}, VGG \citep{simonyan2014very}, and Resnet \citep{he2016deep}) and a set of strategies to exploit the backbone features in the fully convolutional saliency model. These strategies often explore information from shallow and deep layers, providing methods for aggregating or improving the multiple-scale features. Some methods use multiple atrous convolutions with different sampling rates \citep{chen2014semantic, chen2017deeplab}; Zhang \textit{et al.} \citep{zhang2017amulet} propose a generic framework that integrates multiple feature-scales into multiple resolutions; in \citep{zhang2018progressive}, an attention-guided recurrent convolutional network is proposed to integrate multi-level features and reintroduce high-level semantic information from deep layers to shallow layers; Zhang \textit{et al.} \citep{zhang2018bi} presented a bi-direction module to pass information between the shallow and deep layers, and use a pyramid fusing strategy to combine multi-scale information.
    
    The methods described above upsample directly from the low-resolution deep layers back to the full-size input layer, resulting in significant information loss. To reduce the negative impact of drastic upsampling, several methods have adopted the encoder-decoder structure to gradually re-scale the low-resolution features \citep{liu2019simple, tang2018quantized, qin2020u2, qin2019basnet, miao2021automsf}. Each method proposes a different strategy to exploit the multi-scale features in the decoder. As examples, Liu \textit{et al.} \citep{liu2019simple} propose a global guidance module based on pyramid pooling to explicitly deliver object-location information in all feature maps of the decoder; Tang \textit{et al.} \citep{tang2018quantized} proposed stacking U-shaped networks and densely connect their layers, providing a shared-memory implementation strategy to alleviate the memory-heavy requirements. In recent work, Miao \textit{et al.} \citep{miao2021automsf} proposes the \textbf{Auto- Multi-Scale Fusion Network} (\textbf{Auto-MSFNet}), which uses Network Architecture Search (NAS) as inspiration to propose an automatic fusion framework instead of trying to elaborate complex human-described strategies for multi-scale fusion. They introduce a new search cell (FusionCell) that receives information from multi-scale features and uses an attention mechanism to select and fuse the most important information.
    
    Boundary information can be used to achieve better object delineation. Early on, Luo \textit{et al.} \citep{luo2017non} proposed a new loss that heavily penalizes boundary regions for training an adaptation of VGG-16; Li \textit{et al.} \citep{li2018contour} proposes a two-step methodology that extracts contours from an input image and then extrapolates a saliency-map out of the contours;  Su \textit{et al.} \citep{su2019selectivity} proposes a three-stream methodology with a boundary location stream, an interior perception stream, and a boundary/region transition prediction; and Zhao \textit{et al.} \citep{zhao2019egnet} uses an encoder-decoder network to extract multi-scale salient-object features from a backbone, and uses both the high-resolution backbone features and the encoder-decoder object-location information to assist an edge-feature extraction that guides the delineation for the final saliency estimation. Instead of explicitly or exclusively using boundary information, Qin \textit{et al.} \citep{qin2019basnet} proposes the \textbf{Boundary-Aware Segmentation Network} (\textbf{BASNET}) with a hybrid loss to represent differences between the ground-truth and the predictions in a pixel, patch, and map level. The loss is a combination of the Binary-Cross-Entropy (pixel-level) \citep{de2005tutorial}, the Multi-scale Structural Similarity \citep{wang2003multiscale} (patch-level), and the Intersection over Union \citep{mattyus2017deeproadmapper} (map-level). Their strategy consists of a prediction module that outputs seven saliency maps (an upsampled output of each decoder layer) and a residual refinement module that outputs one saliency map at the final decoder layer. All output maps are used when computing the hybrid loss, and both modules are trained in parallel.
    
    All earlier methods require a backbone pretrained in a large classification dataset and often require many training images for fine-tuning. Recently, Qin \textit{et al.} \citep{qin2020u2} proposed a nested U-shaped network by the name of \textbf{U²Net}, that has achieved impressive results without requiring a pretrained backbone. Their approach consists of extracting and fusing multiple-scale information throughout all stages of the encoder-decoder by modeling each stage as a U-shaped network. By doing so, they can extract deep information during all network stages without significantly increasing the number of parameters (due to the downscaling inside each inner U-net).
    
    Even though each presented method has specific characteristics and shortcomings, the main goal of this paper is to aggregate image-intrinsic information to the complex object-inspired models provided by the deep saliency methods. In this regard, we selected as baseline three state-of-the-art saliency estimators that apply different strategies to learn their models: U²Net \citep{qin2020u2}, which has the novelty of not requiring a pre-trained backbone; BASNet \citep{qin2019basnet}, that provides a higher delineation precision due to its boundary-awareness; and Auto-MSFNet \citep{miao2021automsf}, which uses a framework to learn the fusion strategies that humans commonly define.

\section{Proposed approach: Iteractive Saliency Enhancement using Superpixel Similarity (ISESS)}\label{sec:sal-enhance}
\begin{figure*}[t!]
        \centering
             \includegraphics[width=0.9\textwidth]{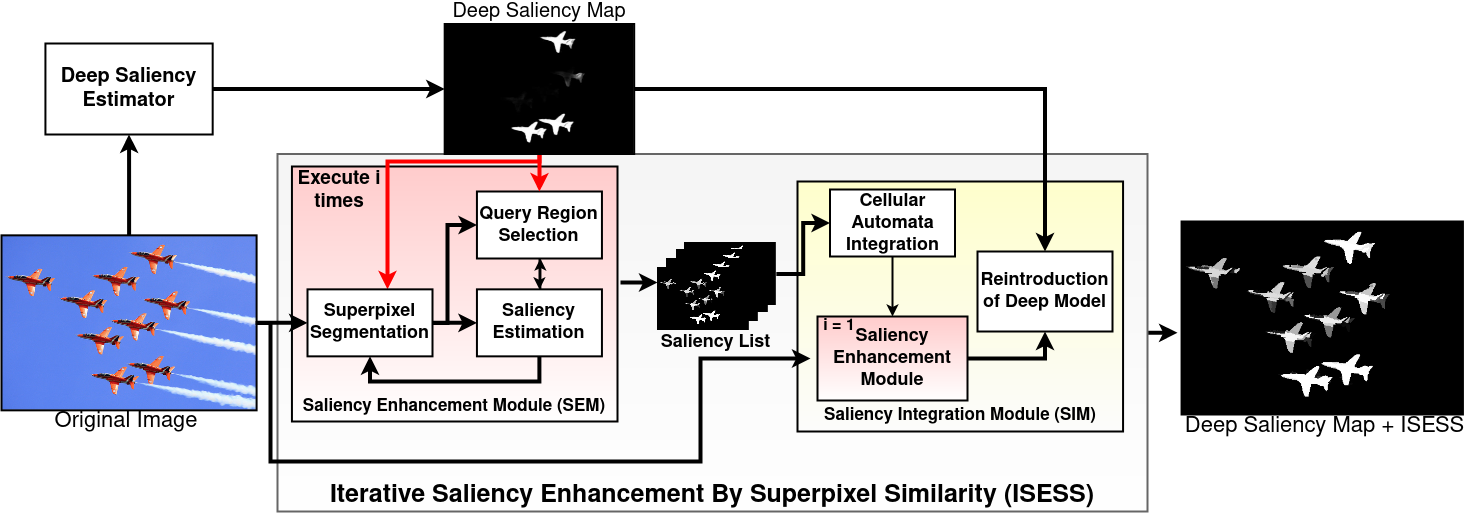} 
        \caption{A depiction of the proposed method. The saliency/superpixel enhancement loop is presented inside the Saliency Enhancement Module (SEM). Note that the SEM inside the Saliency Integration Module (SIM) runs only once. The red arrows represent a flow happening only on the first iteration. }
        \label{fig:sess-diagram}
\end{figure*}

ISESS aims to improve the results of complex deep-learning-based SOD models by adding image-intrinsic information. For such, ISESS combines alternate executions of object-based superpixel segmentation and superpixel-based saliency estimation. 
The multiple executions of both operations generate enhanced saliency maps for integration into a post-processed map, which is subsequently merged with the initial saliency map of the deep SOD model. Figure \ref{fig:sess-diagram} illustrates the whole process, described in Sections \ref{subsec:sem} and \ref{subsec:sim}. 

\subsection{Saliency Enhancement Module}\label{subsec:sem}

The saliency enhancement module starts by computing a superpixel segmentation since ISESS is a superpixel-based saliency enhancer. For such, we use the Object-based Iterative Spanning Forest algorithm (OISF) \citep{belem2019oisf}. 

In short, OISF~\footnote{The OISF method is available at \url{https://github.com/LIDS-UNICAMP/OISF}.} represents an image as a graph, whose pixels are the nodes and the arcs connect 8-adjacent pixels, elects seed pixels based on an input object saliency map, and executes the Image Foresting Transform (IFT) algorithm~\citep{falcao2004ift} followed by seed refinement multiple times to obtain a final superpixel segmentation. Each superpixel is then represented as an optimum-path tree rooted at its most closely connected seed. 

For the initial seed sampling based on a saliency map, we use OSMOX~\citep{belem2019importance}, which allows user control over the ratio of seeds to be placed inside and outside the salient objects with no need to binarize the saliency map. For seed refinement, we set the new seed of each superpixel to be the closest pixel to its geometrical center.

For this setup of OISF, the user has to provide four parameters: $\alpha$ to control the importance of superpixel regularity; $\beta$ to control the importance of boundary adherence; $\gamma$ to provide a balance between saliency information and pixel colors; and $i_o$ the number of iterations to obtain a final superpixel segmentation. Within the method proposed in this paper, we fixed $\alpha = 12$ and $\beta = 0.5$ as suggested in ~\citep{vargas2018isf}, while $\gamma$ and $i_o$ were tuned by grid searching (see Section \ref{subsec:setup}).

The only specific implementation change regarding the superpixel segmentation algorithm is the percentage of object seeds required by the seed sampler. Instead of fixing a percentage that would fit most images, we set the number of seeds inside salient areas to be $n_{os} = n_s*n_c$, where $n_c$ is the number of connected components found by Otsu binarization of the saliency map resultant from the last iteration, and $n_s$ is a parameter. This way, the number of superpixels depends on how many salient objects the oversegmentation is trying to represent.

Within this paper, let the saliency score of either a pixel or a superpixel be represented as $\textbf{s}(\cdot)$. At the end of each iteration, the saliency score of a superpixel is given by the previously computed pixel-wise saliency map and is taken as the mean saliency score of all pixels inside the superpixel. We do not reuse the saliency of a superpixel directly because the superpixel segmentation changes at each iteration. 

Let $\mathcal{S}$ be a superset of all superpixels, $\textbf{F} : \mathcal{S}\rightarrow \mathbb{R}^{m}$ map to every superpixel the mean feature vector of all its pixels --- in this paper, $m = 3$ because we use the pixel colors in the CIELab color-space as features. We define two query lists,  $\mathcal{Q}_F$ for foreground superpixels, and $\mathcal{Q}_B$ for background ones, where $\mathcal{Q}_F \cup \mathcal{Q}_B = \mathcal{S}$. Take $\psi$ to be the Otsu threshold of the previously computed saliency map, then, the query lists are defined as follows: $\textit{S} \in \mathcal{Q}_F \leftrightarrow \textbf{s}(\textit{S}) \geq \psi$, and similarly, $\textit{S} \in \mathcal{Q}_B \leftrightarrow \textbf{s}(\textit{S}) < \psi$.

%Let $\mathcal{S}$ be a superset of all superpixels, $\textbf{F} : \mathcal{S}\rightarrow \mathbb{R}^{3}$ map to every superpixel the mean color of all its pixels in the CIELab color-space. We define two query lists,  $\mathcal{Q}_F$ for foreground superpixels, and $\mathcal{Q}_B$ for background ones, where $\mathcal{Q}_F \cup \mathcal{Q}_B = \mathcal{S}$. Take $\psi$ to be the Otsu threshold of the previously computed saliency map, then, the query lists are defined as follows: $\textit{S} \in \mathcal{Q}_F \leftrightarrow \textbf{s}(\textit{S}) \geq \psi$, and similarly, $\textit{S} \in \mathcal{Q}_B \leftrightarrow \textbf{s}(\textit{S}) < \psi$.

Taking $\textit{S}, \textit{R} \in \mathcal{S}$ as superpixels, we define a Gaussian-weighted similarity measure between superpixels as: 

\begin{equation}
    \textbf{sim}(\textit{S}, \textit{R}) = \exp^{\frac{-\textbf{d}(\textbf{F}(\textit{S}), \textbf{F}(\textit{R}))}{\sigma^2}},
\end{equation}

where $\textbf{d}(\textbf{F}(\textit{S}), \textbf{F}(\textit{R}))$ is the euclidean distance between the superpixel's mean features, and $\sigma^2 = 0.01$ is the variance used to regulate the dispersion of similarity values. 

Using the similarity measure and the query lists, we define two saliency scores for each superpixel: one based on foreground queries, and the other on background ones. For the foreground saliency score, a region is deemed to be salient if it shares similar characteristics to at least one other foreground region:

\begin{equation}
     \textbf{s}_f(\textit{S}) = \max\limits_{\forall \textit{R} \in \mathbb{Q}_F}\{{\textbf{\textbf{sim}(\textit{S}, \textit{R})}}\}
     \label{eq:foreground-sal}
\end{equation}

However, when $\textit{S}=\textit{S}_f \in \mathbb{Q}_F$ is a foreground query, its foreground saliency score equals one, which is an out-scaled value as compared to other gaussian-weighted similarities. To keep foreground queries with the highest score but in the same scale of the remaining superpixels' scores, we update their values to match the highest non-foreground query region's saliency: $\textbf{s}_f(\textit{S}_f) = \max\limits_{\forall \textit{Q} \notin \mathcal{Q}_F}\{{\textbf{s}_f(\textit{Q})}\}$. By doing so, we allow for a forced exploration of possible foreground regions at each iteration.

Similarly, the background-updated saliency score defines that a superpixel is salient if it has low similarity to most background queries:

\begin{equation}
     \textbf{s}_b(\textit{S}) = 1 - \frac{\sum\limits_{\forall \textit{R} \in \mathbb{Q}_B, \textit{S} \neq \textit{R}}{\textbf{sim(\textit{S}, \textit{R})}}}{|\mathbb{Q}_B|}.
\end{equation}

\begin{figure}[t!]
    \centering
    \begin{tabular}{c c}
             \includegraphics[width=0.2\textwidth]{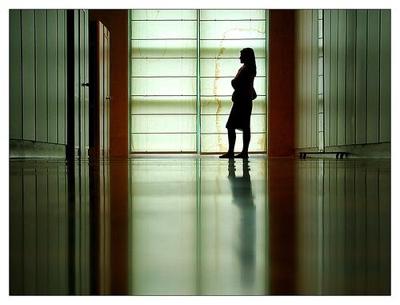} &
             \includegraphics[width=0.2\textwidth]{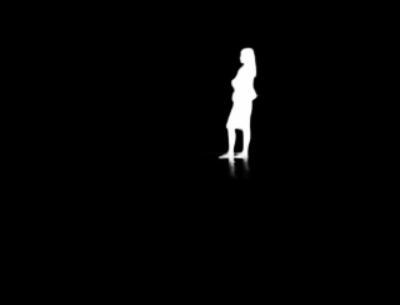} \\
             (a) & (b) \\
             \includegraphics[width=0.2\textwidth]{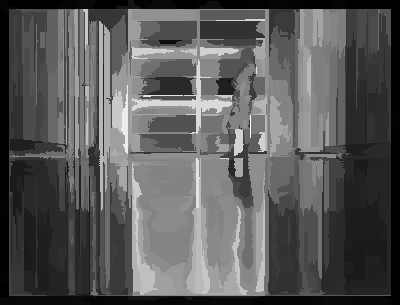} &
             \includegraphics[width=0.2\textwidth]{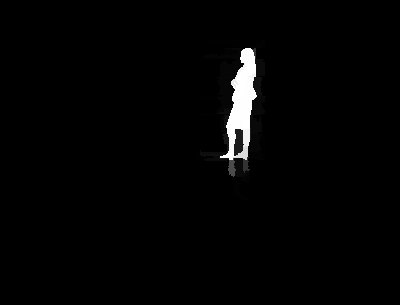} \\
              (c) & (d) 
        \end{tabular}
    \caption{Background-based saliency with $d_{0.5} < 0.1$ due to a complex background, resulting on a map with little relevant information. (a) original image; (b) U²Net map; (c) background-based saliency; (d) saliency used instead of (c).}
    \label{fig:background-sal}
\end{figure}

When the background is too complex, the normalized mean difference of all superpixels will be close to fifty percent, resulting in a mostly gray map with little aggregated information (Figure \ref{fig:background-sal}). To detect those scenarios, we compute a map-wise distance to fifty percent probability $d_{0.5} = \frac{1}{|\mathbb{S}|} \sum\limits_{\forall \textit{S} \in \mathbb{S}}{(\textbf{s}_b(\textit{S}) - 0.5)^2 |\textit{S}|}$. For maps with $d_{0.5} < 0.1$, we set the background saliency score to be the saliency of the previous iteration $\textbf{s}_b(\textit{S}) = \textbf{s}^{i-1}(\textit{S})$, where $\textbf{s}^{i-1}$ is the final saliency computed in the last iteration and $\textbf{s}^{0}$ is the initial deep-learning saliency.

The final saliency score for the iteration is the product of both saliencies:

\begin{equation}
     \textbf{s}^i(\textit{S}) = \textbf{s}_f(\textit{S}) * \textbf{s}_b(\textit{S}).
\end{equation}

By multiplying both scores, the combined saliency ignores any region that was hardly taken as non-salient by foreground or background saliency and keeps a high score for regions taken as salient in both maps. The new score is saved together with the superpixel representation used to create it. Then, the saliency is fed to the superpixel algorithm to restart the enhancement cycle. We reduce the number of superpixels by $20\%$ at each iteration to a minimum of $200$ superpixels.

The result of all iterations is then combined using the method described in Section \ref{subsec:sim}

\subsection{Saliency Integration Module}\label{subsec:sim}
    To integrate the saliency maps created throughout the iterations, we use a traditional unsupervised saliency integrator method \citep{qin2015saliency} that models an update function using a Bayesian framework to iteratively update a Cellular Automata whose cells are derived from a stack of the saliency maps. Note that we are using saliency integration to combine the multiple outputs of our iterative approach, not aggregating different saliency methods.
    
    In short, the saliency maps are stacked to form a three-dimensional grid where the (x,y) coordinates match the saliency map's (x,y) coordinates and z is the direction in which the different maps are stacked. The pixels are then updated over $\textit{it}$ iterations in a way that the new pixel saliency depends on how consistently salient its adjacency is. The adjacency used is a cuboid adjacency $\mathbb{C}_p$ around the pixel $p$, so that a pixel $q$ is considered adjacent to a $p$ if $|(x_p - x_q)| \leq 1, |(y_p - y_q)| \leq 1$; \textit{i.e.} a 4-adjacency extended to all maps on the z-axis.
    
    For the update rule, the saliency score is represented as log odds: $\textbf{l}(p) = \frac{\textbf{s}(p)}{1 - \textbf{s}(\textit{S})}$ so that the value can be updated by performing subsequent sums. The value changes according to the rule:
\begin{equation}
     \textbf{s}^{t+1}(p) = \textbf{l}(p) + \sum\limits_{\forall q \in \mathbb{C}_p}{\lambda\delta(q)},
\end{equation}

where $\lambda$ is a constant that regulates the update rate, and $\delta(\cdot) \in \{-1, 1\}$ is a signal function that makes the saliency either increase or decrease at each update depending on whether $\textbf{s}(q)$ is greater or smaller than the Otsu threshold of the saliency map that originated the cell. Essentially, if a pixel is surrounded by more foreground than background regions, its saliency increase over time by a rate of $\lambda$; likewise, the saliency decreases for pixels in a non-salient adjacency.

    Afterward, the integrated map is run through the last iteration of saliency estimation using Equation \ref{eq:foreground-sal} and the initial number of superpixels, creating the last color-based saliency score $\textbf{s}_c$. This last estimation serves two purposes: improve object delineation by relying on the quality of the superpixel algorithm --- rather than depending on the cuboid adjacency relation; and eliminate unwanted superpixel leakage that may occur when the superpixel number is reduced (Section \ref{subsec:number-of-superpixel-increase}). 
    
    Lastly, we reintroduce the deep object-inspired model by averaging its saliency values inside the superpixels of the last segmentation, which defines another saliency score for each superpixel:
\begin{equation}
     \textbf{s}_d(\textit{S}) = \frac{1}{|\textit{S}|}\sum\limits_{\forall p \in \textit{S}}{\textbf{s}_0(p)},
\end{equation}

where $\textbf{s}_0(p)$ is the saliency score provided by the network. The final saliency score is than defined as:

\begin{equation}
     \textbf{s}_f(p) = \max\{\textbf{s}_d(p), \textbf{s}^{i}(p)\}.
\end{equation}

Therefore, the final saliency map will highlight salient regions according to the image-intrinsic or the superpixel-delineated object-inspired model.
\section{Experiments and Results}\label{sec:results}

\subsection{Datasets and experimental setup}\label{subsec:setup}
    \textbf{Datasets:} We used five well-known datasets for comparison among SOD methods.
    \textbf{DUT\_OMRON} is composed of 5168 images containing one or two complex foreground objects in a relatively cluttered background; \textbf{HKU-IS} contains 4447 images with one or multiple low contrast foreground objects each; \textbf{ECSSD} consists of 1000 images with mostly one large salient object per image in a complex background; \textbf{ICoSeg} consists of  643 images usually with several foreground objects each; and \textbf{SED2} is formed by 100 images with two foreground objects per image.
    
    \textbf{Parameter tuning:} The baselines networks were used as provided by the authors since they have been pretrained on datasets with similar images (e.g., \textbf{DUT\_OMRON}). For ISESS, we randomly selected 50 images from each dataset, totalizing a training set with 300 images, to optimize its parameters by grid search. The remaining images compose the test sets of each dataset. Grid search was performed with saliency maps from each baseline, creating three sets of parameters (one per baseline). Most parameters were optimized to the same value independently of the baseline, except for the number of iterations $i$, the number of superpixels $n$, the number of foreground seeds per component $n_s$, and the number of OISF iterations $i_o$. Taking the order U²Net, BASNET, and MSFNet, the parameters were respectively, $i = \{12, 9, 12\}$, $n = \{2500, 200, 2500\}$, $n_s = \{10, 30, 30\}$, and $i_o = \{5, 3, 1\}$. The rest of the parameters with fixed values for all methods were optimized to be: $\gamma = 10.0$, $\sigma^2 = 0.01$, $\lambda = 0.0001$, $t = 3$, where $\gamma$ is related to the superpixel segmentation algorithm, $\sigma^2$ is the gaussian variance for the graph node similarity, $\lambda$ and $t$ are related to the cuboid integration.
    
    \textbf{Evaluation metrics: } We used six measures for quantitative assessment. \textbf{(1) Mean Structural-measure}, $S_m$, which evaluates the structural similarity between a saliency map and its ground-truth; \textbf{(2) max F-measure},  $\max F_\beta$, which represents a balanced evaluation of precision and recall for simple threshold segmentation of the saliency maps; \textbf{(3) weighted F-measure},  $F_\beta^w$, which is similar to $\max F_\beta$ but keeps most saliency nuances by not requiring a binary mask; \textbf{(4) Mean Absolute Error}, $MAE$, that provides a pixel-wise difference between the output map and expected segmentation; \textbf{(5) mean Enhanced-alignment measure}, $E_\psi^m$, used to evaluate local and global similarities simultaneously; and \textbf{(6) precision-recall curves}, which display precision-recall values between the expected segmentation and saliency maps binarized by varying thresholds.
    
\begin{figure}[t!]
    \centering
        \begin{tabular}{c c}
                 \includegraphics[width=0.2\textwidth]{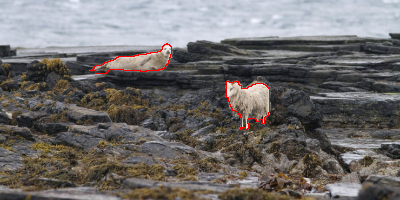} &
                 \includegraphics[width=0.2\textwidth]{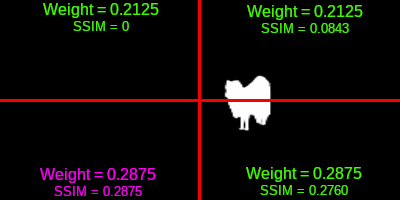} \\
                 (a) & (b)\\
                 \includegraphics[width=0.2\textwidth]{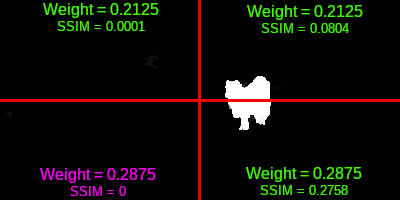} &
                 \includegraphics[width=0.2\textwidth]{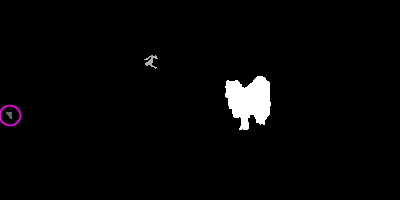} \\
                 (c) & (d) \\
            \end{tabular}
        \caption{Example of drastic $S_m$ difference between two similar saliency maps. (a) original image with ground-truth objects delineated in red; (b) BASNET saliency map divided into quadrants, with each quadrant weight and Structural Similarity values annotated; (c) same as (b) but with the map enhanced by ISESS; (d) histogram-manipulated (c) to show the detail that caused the large difference in similarity value.}
        \label{fig:smeasure-problems}
\end{figure}

    \textbf{Post processing: } ISESS can sometimes score a superpixel with close-to-zero saliency values. These small saliencies are not visually perceived by the observer but can significantly impair the presented metrics (Figure \ref{fig:smeasure-problems}). In the image presented, the bottom-left quadrants (with a little more than a fourth of the total image weight) had a superpixel slightly salient (a value below $2\%$ of the maximum saliency) by ISESS, which caused the local SSIM to change from a full match on the original map to a complete miss on the enhanced map. To deal with these small values, we eliminate regions in the map with saliency lower than half of the Otsu threshold. 
    
\subsection{Ablation study}\label{subsec:ablation}
    To analyze the impact of some decisions within our proposal, we present ablation studies on (i) the effects of re-introducing the deep model at the end of the last iteration; (ii) re-using the initial number of superpixels at the last iteration. We used the same parameters as the ones reported in Section \ref{subsec:setup} for the ablation studies.
    
\subsubsection{Ablation on reintroducing the deep saliency model}\label{subsec:reintroducing-deep-model}
    The goal of ISESS is to improve the complex and robust deep models by adding information more closely related to what a human observer expects. Not re-introducing the deep model would imply creating a color-based model that keeps all the robustness of the deep-learning ones, which is hardly feasible. 
    
\begin{figure}[t!]
    \centering
        \begin{tabular}{c c}
                 \includegraphics[width=0.18\textwidth]{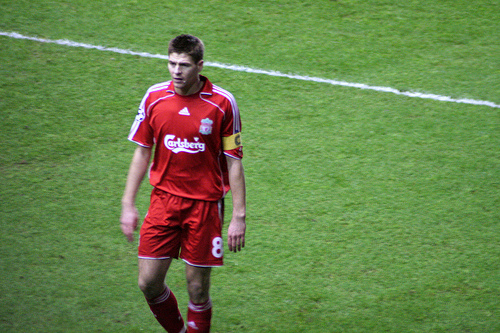} &
                 \includegraphics[width=0.18\textwidth]{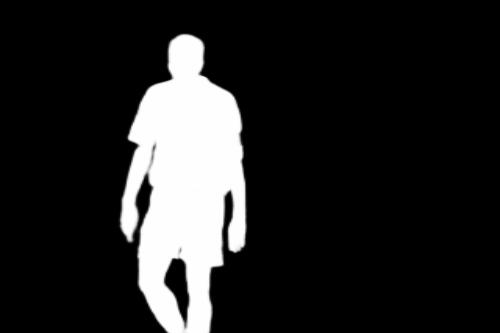} \\
                 (a) & (b) \\
                 \includegraphics[width=0.18\textwidth]{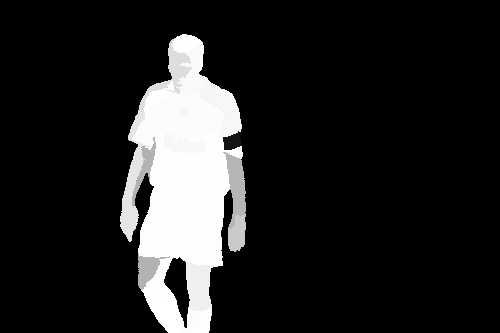} &
                 \includegraphics[width=0.18\textwidth]{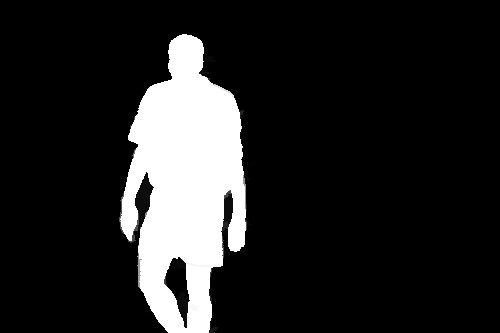} \\
                 (c) & (d) 
            \end{tabular}
        \caption{Good example of the re-introduction of the deep-model saliency. (a) original image; (b) U²Net saliency map; (c-d) ISESS without and with the reintroduction of the deep model, respectively}
        \label{fig:deep-reintro-good}
\end{figure}

    An example of why the proposed color-based model is often insufficient can be seen in Figure \ref{fig:deep-reintro-good}. The object of interest was less uniformly salient: The armband the player is wearing has significantly different colors than the rest of the object, which caused the model not to consider it as part of the foreground. In that example, ISESS mainly contributed to creating sharper edges.
    
\begin{figure}[t!]
    \centering
        \begin{tabular}{c c c c}
                 \includegraphics[width=0.095\textwidth]{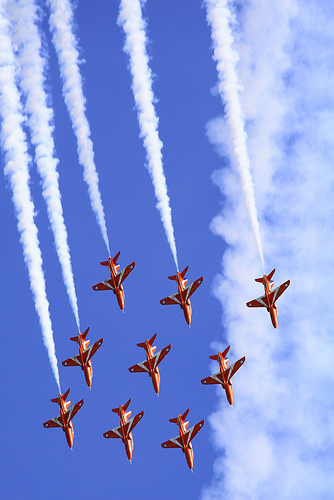} &
                 \includegraphics[width=0.095\textwidth]{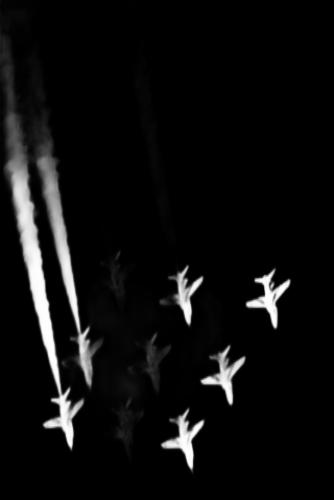} &
                 \includegraphics[width=0.095\textwidth]{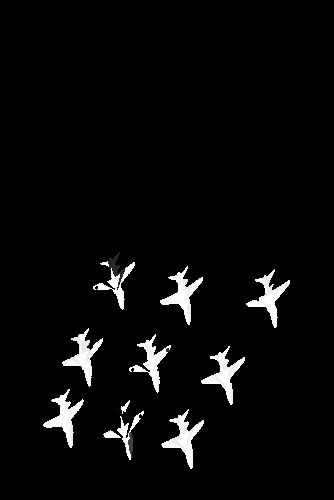} &
                 \includegraphics[width=0.095\textwidth]{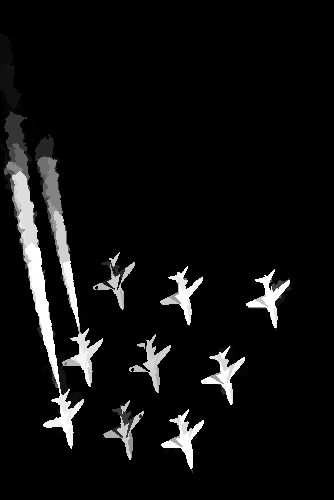} \\
                 (a) & (b) & (c) & (d)
            \end{tabular}
        \caption{Bad example of the re-introduction of the deep-model saliency. (a) original image; (b) U²Net saliency map; (c-d) ISESS without and with the reintroduction of the deep model, respectively}
        \label{fig:deep-reintro-bad}
\end{figure}

    The downside of re-introducing the deep model is that it reduces the influence of ISESS on wrongfully salient regions. Take Figure \ref{fig:deep-reintro-bad} as an example: The deep-model wrongfully highlighted part of the airplanes' smoke trail; ISESS did perceive this to be a mistake and removed it entirely; the re-introduction of the deep-model also brought back the error. However, the smoke trail's saliency was reduced thanks to combining the deep model with the superpixel segmentation.
    
\subsubsection{Ablation on using the initial number of superpixels for the last iteration}\label{subsec:number-of-superpixel-increase}
\begin{figure}[t!]
    \centering
        \begin{tabular}{c c c}
                 \includegraphics[width=0.135\textwidth]{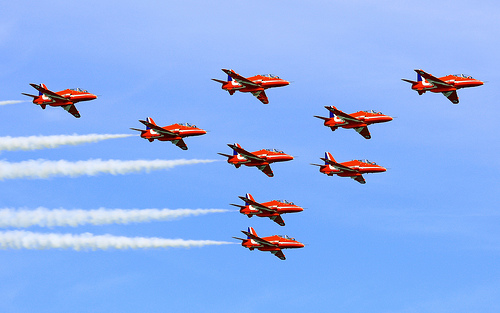} &
                 \includegraphics[width=0.135\textwidth]{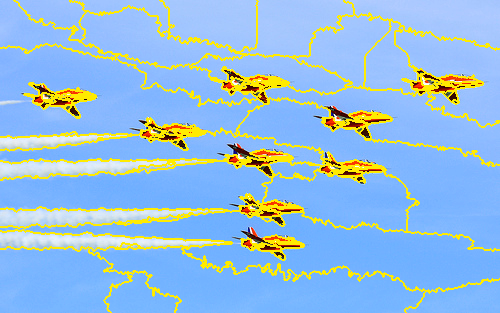} &
                 \includegraphics[width=0.135\textwidth]{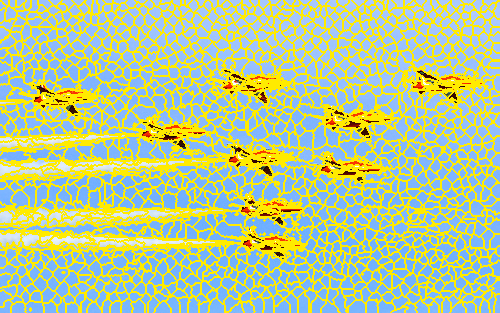} \\
                 (a) & (b) & (c)\\
                 \includegraphics[width=0.135\textwidth]{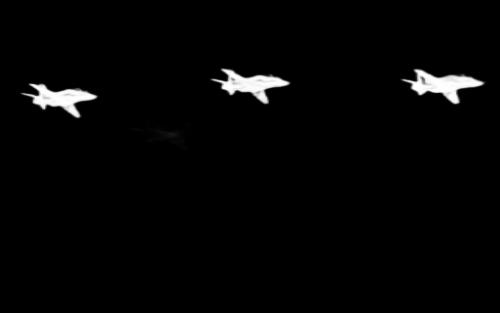} &
                 \includegraphics[width=0.135\textwidth]{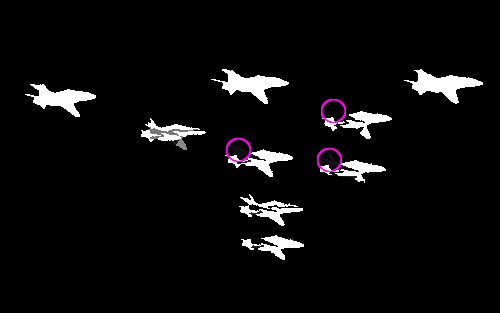} &
                 \includegraphics[width=0.135\textwidth]{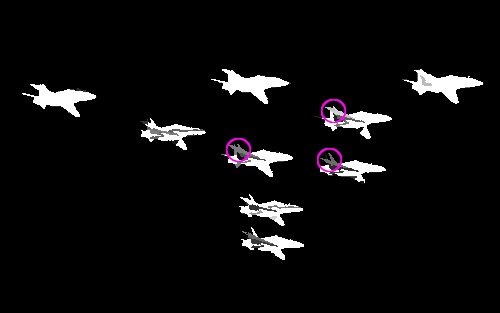} \\
                 (d) & (e) & (f)
            \end{tabular}
        \caption{Bad object representation due to the superpixel decrease. (a-c) original image and superpixel segmentation using the reduced and the full number of superpixels, respectively; (d-f) U²Net, U²Net+ISESS without and with the superpixel number increase at the last iteration, respectively. Note part of the object missing (pink circle) due to superpixel leakege.}
        \label{fig:superpixel-leakage}
\end{figure}

    As the iterations progress and the number of superpixels decreases, the object-representation provided by the superpixels can worsen due to either leakage to the background or under-segmentation of too distinct object parts. As exemplified in Figure \ref{fig:superpixel-leakage}, the wings of the airplanes were lost in the final saliency map due to superpixel leakage. The model could adequately highlight most of the planes' wings by increasing the number of superpixels at the last iteration.
    
\subsection{Quantitative comparisons}\label{subsec:quantitative-comparison}
    Table \ref{tab:quantitative-results} shows that the proposed method can consistently improve $max F_\beta$, $F_\beta^w$, $MAE$ and $E_\psi^m$ for U²Net \citep{qin2020u2} and BASNET \citep{qin2019basnet} in all five datasets, and achieve similar $S_m$. On both datasets composed mostly of images containing more than one object (Sed2 and ICoSeg), the results were even more expressive, indicating that ISESS can considerably improve the saliency representation of similar objects.
    
\begin{figure}[t!]
    \centering
    \begin{tabular}{c c c c}
             \includegraphics[width=0.085\textwidth]{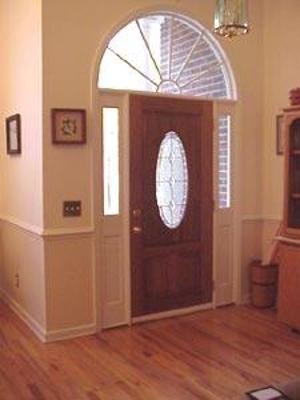} &
             \includegraphics[width=0.085\textwidth]{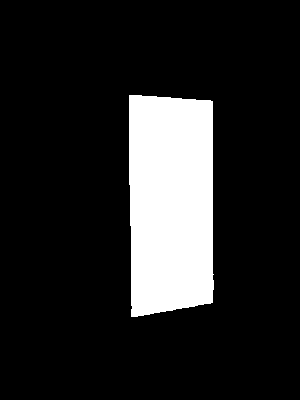} &
             \includegraphics[width=0.085\textwidth]{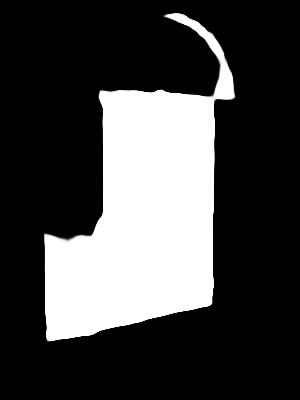} &
             \includegraphics[width=0.085\textwidth]{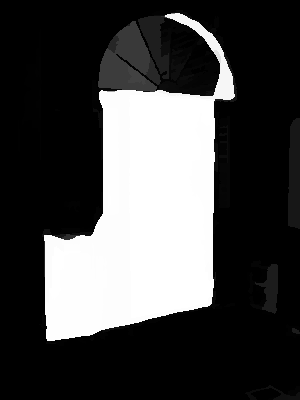} \\
             \includegraphics[width=0.085\textwidth]{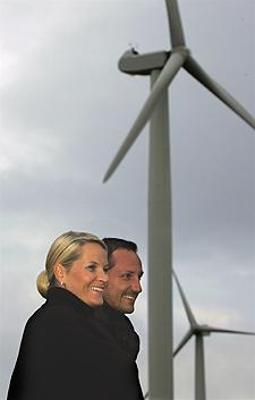} &
             \includegraphics[width=0.085\textwidth]{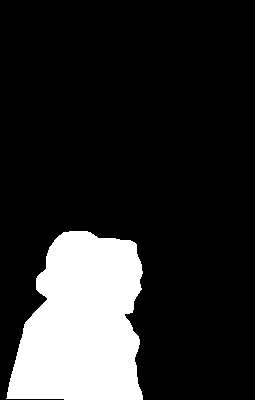} &
             \includegraphics[width=0.085\textwidth]{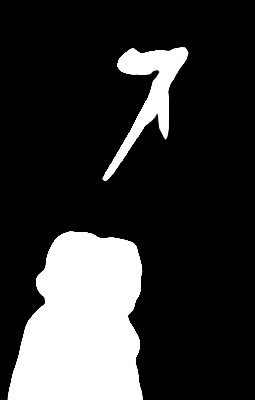} &
             \includegraphics[width=0.085\textwidth]{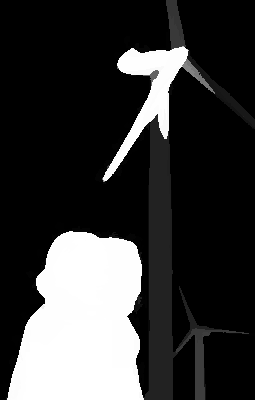} \\
              (a) & (b) & (c) & (d) \\
        \end{tabular}
    \caption{ISESS improving the saliency of wrogfully salient object partially highlighted by Auto-MSFNet. (a) original image; (b) label; (c) saliency map by Auto-MSFNet; (d) (c) enhanced by the proposed ISESS.}
    \label{fig:msf-problems}
\end{figure}

    The precision-recall curves show a clear advantage of ISESS-enhanced maps over the non-enhanced ones on datasets mainly consisting of multiple objects (sed2, and icoseg). There is a breakpoint where ISESS loses precision more rapidly than non-enhanced maps on the other three datasets. We attribute the rapid precision loss to ISESS highlighting more parts of non-salient objects deemed partially salient by the deep models (Figure \ref{fig:msf-problems}). By looking at the $maxF_\beta$, we see that by segmenting ISESS maps using an adequate threshold, the segmentation results are often better than the non-enhanced maps on almost every combination of method and dataset. 
    
    The better representation of non-salient objects discussed before also highly impacts the $S_m$. Similar to the example shown in Figure \ref{fig:smeasure-problems}, spatially extending the saliency of a non-salient object to other image quadrants drastically affects the quality of the map according to the Mean Structural-measure.
    
    Regarding Auto-MSFNet, apart from the ICoSeg and SED2, ISESS could not improve its metrics. By analyzing the precision-recall curves, we identified that Auto-MSFNet has the lowest precision of all three methods. We realized that the AutoMSF-Net creates more frequently partially salient regions on non-salient objects. Both examples used in Figure \ref{fig:msf-problems} were taken from the AutoMSF-Net results. Therefore, ISESS seems to over-trust the deep model on non-salient regions.
    
\begin{figure*}[t!]
        \centering
        \begin{tabular}{c c c}
                 \includegraphics[width=0.28\textwidth]{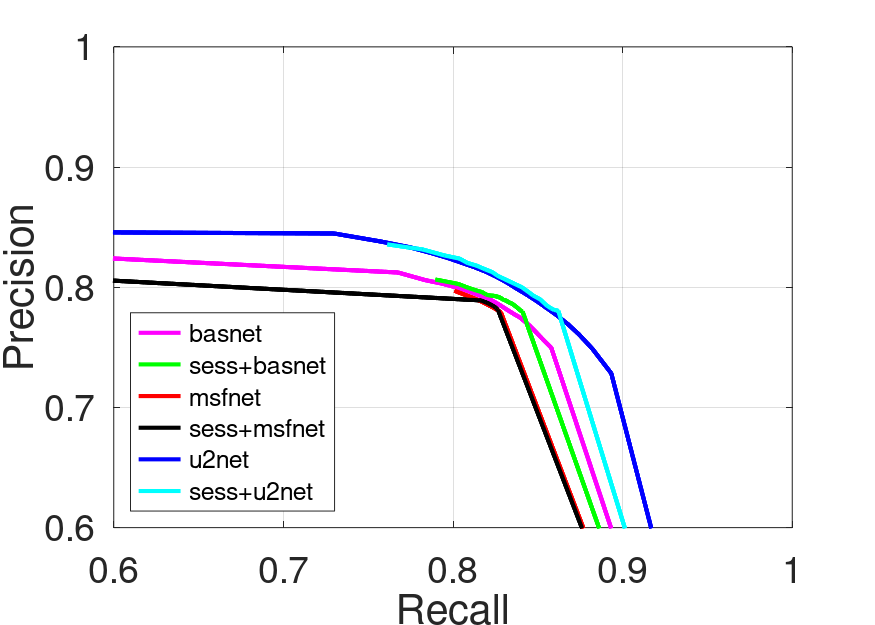} &
                 \includegraphics[width=0.28\textwidth]{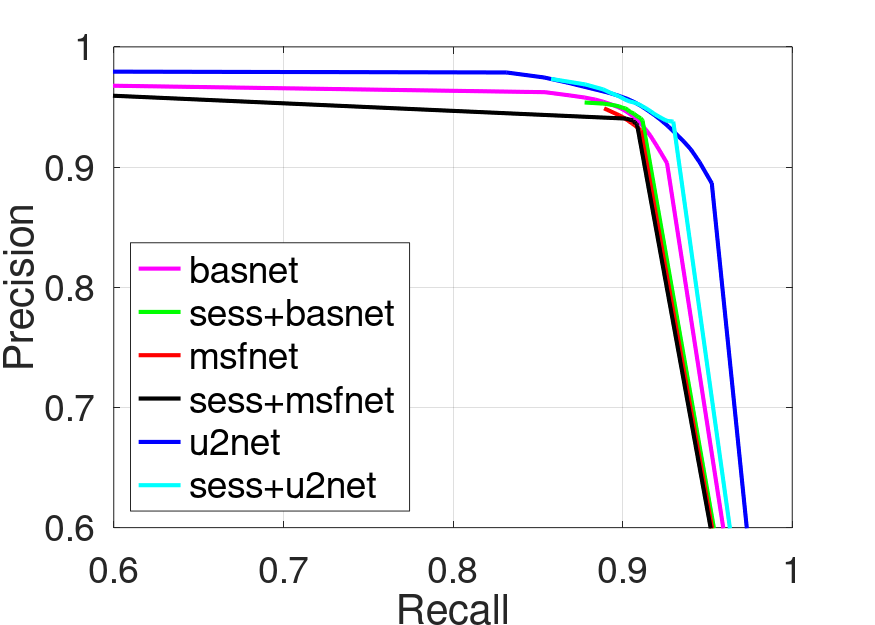} &
                 \includegraphics[width=0.28\textwidth]{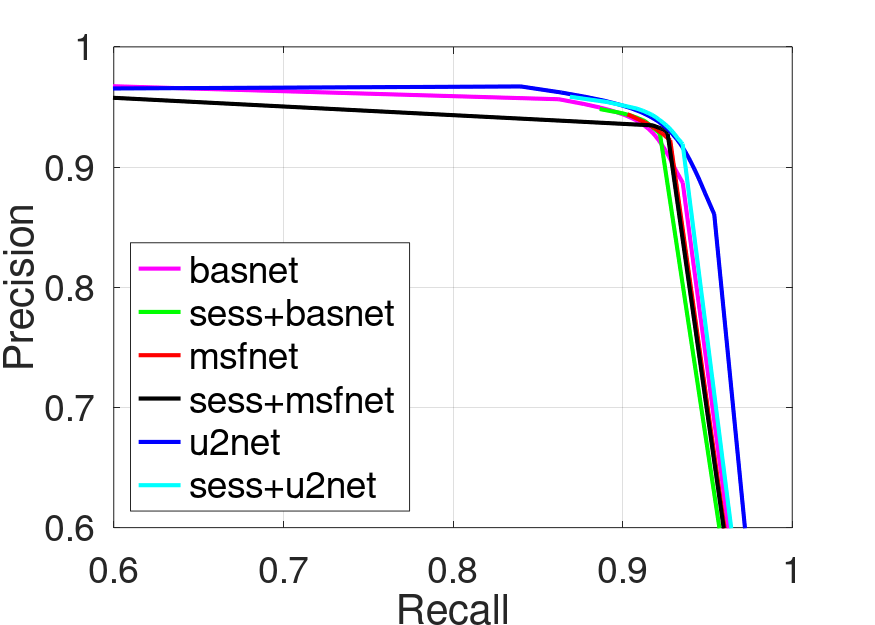} \\
                 (a) & (b) & (c)\\
            \end{tabular}
            
        \begin{tabular}{c c}
                 \includegraphics[width=0.3\textwidth]{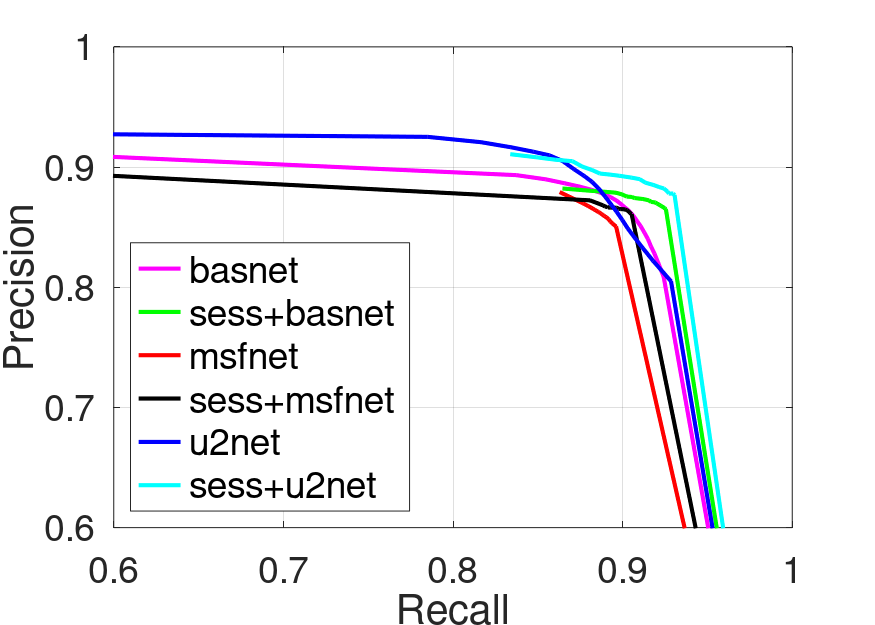} &
                 \includegraphics[width=0.3\textwidth]{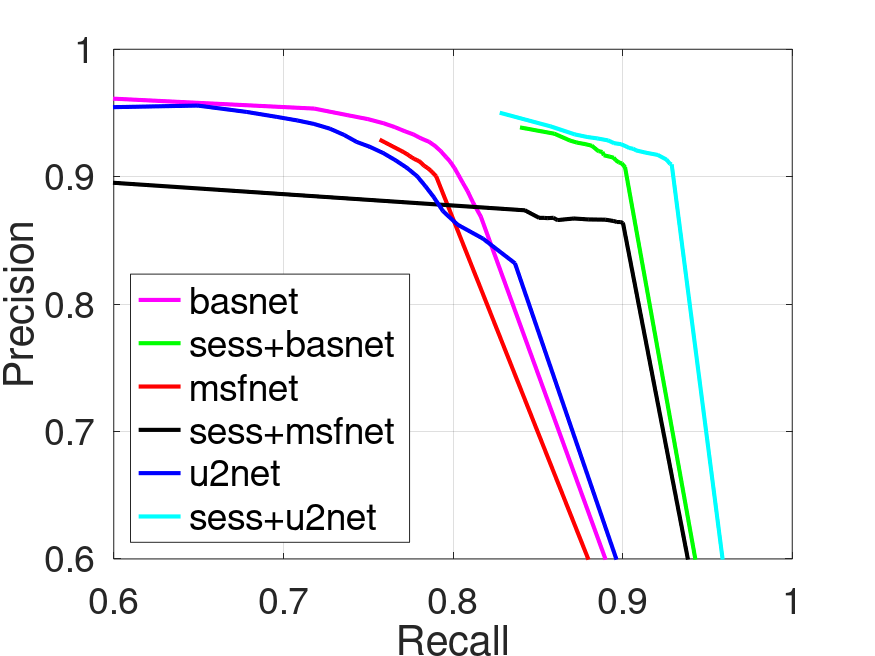}  \\
                 (d) & (f) \\
            \end{tabular}
        \caption{The Precision-Recall curves of all methods with and without enhancement in all datasets. (a) DUT\_OMRON; (b) ECSSD; (c) HKU-IS; (d) ICOSEG; (e) SED2.}
        \label{fig:pr-curves}
\end{figure*}

\begin{table}[htb]
 \centering
  \rowcolors{2}{gray!25}{white}
    \resizebox{.45\textwidth}{!}{%
    \begin{tabular}{l|l|l|l|l|l}
    \hline
    \rowcolor{white}
    \textcolor{red}{\textbf{Sed2}}          & $S_m$ $\uparrow$          & $max F_\beta$ $\uparrow$  & $F_\beta^w$ $\uparrow$    & $MAE$ $\downarrow$        & $E_\psi^m$ $\uparrow$     \\ \hline
    U²Net                                   & 0.8193                    & 0.8140                    & 0.7704                    & 0.0558                    & 0.8529                    \\ \hline
    \textbf{U²Net+ISESS}                     & \textcolor{blue}{0.8527}  & \textcolor{blue}{0.8404}  & \textcolor{blue}{0.8087}  & \textcolor{blue}{0.0483}  & \textcolor{blue}{0.8816}  \\ \hline
    BASNET                                  & 0.8558                    & 0.8565                    & 0.8119                    & 0.0514                    & 0.8881                    \\ \hline
    \textbf{BASNET+ISESS}                    & \textcolor{blue}{0.8643}  & \textcolor{blue}{0.8647}  & \textcolor{blue}{0.8375}  & \textcolor{blue}{0.0470}  & \textcolor{blue}{0.9037}  \\ \hline
    Auto-MSFNet                                  & 0.8374                    & 0.8333                    & 0.8032                    & 0.0544                    & 0.8800                    \\ \hline
    \textbf{Auto-MSFNet+ISESS}                    & \textcolor{blue}{0.8378}  & \textcolor{blue}{0.8518}  & \textcolor{blue}{0.8182}  & \textcolor{blue}{0.0524}  & \textcolor{blue}{0.8881}  \\ 
    \hline\hline
    \textcolor{red}{\textbf{ECSSD}}         & $S_m$ $\uparrow$          & $max F_\beta$ $\uparrow$  & $F_\beta^w$ $\uparrow$    & $MAE$ $\downarrow$        & $E_\psi^m$ $\uparrow$     \\ \hline
    \rowcolor{white}
    U²Net                                   & \textcolor{blue}{0.9209}  & 0.9286                    & 0.8993                    & 0.0370                    & 0.9407                    \\ \hline
    \rowcolor{gray!25}
    \textbf{U²Net+ISESS}                     & 0.9182                    & \textcolor{blue}{0.9297}  & \textcolor{blue}{0.9099}  & \textcolor{blue}{0.0339}  & \textcolor{blue}{0.9467}  \\ \hline
    \rowcolor{white}
    BASNET                                  & \textcolor{blue}{0.9142}  & 0.9262  & 0.8992                    & 0.0384                    & 0.9414                    \\ \hline
    \rowcolor{gray!25}
    \textbf{BASNET+ISESS}                    & 0.9123                    & \textcolor{blue}{0.9267}            & \textcolor{blue}{0.9075}  & \textcolor{blue}{0.0355}  & \textcolor{blue}{0.9468}  \\ \hline
    \rowcolor{white}
    Auto-MSFNet                                  & 0.9067                    & \textcolor{blue}{0.9229}  & \textcolor{blue}{0.9032}  & \textcolor{blue}{0.0387}  & \textcolor{blue}{0.9427}  \\ \hline
    \rowcolor{gray!25}
    \textbf{Auto-MSFNet+ISESS}                    & \textcolor{blue}{0.9069}  & 0.9203                    & 0.9026                    & 0.0389                    & 0.9411                    \\ \hline\hline
    
    \rowcolor{white}
    \textcolor{red}{\textbf{DUT\_OMRON}}    & $S_m$ $\uparrow$          & $max F_\beta$ $\uparrow$  & $F_\beta^w$ $\uparrow$    & $MAE$ $\downarrow$        & $E_\psi^m$ $\uparrow$     \\ \hline
    U²Net                                   & \textcolor{blue}{0.8444}  & 0.7868                    & 0.7491                    & 0.0541                    & 0.8627                    \\ \hline
    \textbf{U²Net+ISESS}                     & 0.8418                    & \textcolor{blue}{0.7895}  & \textcolor{blue}{0.7647}  & \textcolor{blue}{0.0518}  & \textcolor{blue}{0.8736}  \\ \hline
    BASNET                                  & \textcolor{blue}{0.8362}  & 0.7750                    & 0.7465                    & 0.0558                    & 0.8607                    \\ \hline
    \textbf{BASNET+ISESS}                    & 0.8335                    & \textcolor{blue}{0.7758}  & \textcolor{blue}{0.7556}  & \textcolor{blue}{0.0548}  & \textcolor{blue}{0.8686}  \\ \hline
    Auto-MSFNet                                  & 0.8321                    & \textcolor{blue}{0.7748}  & \textcolor{blue}{0.7536}  & \textcolor{blue}{0.049}   & \textcolor{blue}{0.8692}  \\ \hline
    \textbf{Auto-MSFNet+ISESS}                    & \textcolor{blue}{0.8323}  & 0.7730                    & 0.7533                    & 0.0491                    & 0.8680                    \\ \hline\hline
    
    \textcolor{red}{\textbf{ICoSeg}}        & $S_m$ $\uparrow$          & $max F_\beta$ $\uparrow$  & $F_\beta^w$ $\uparrow$    & $MAE$ $\downarrow$        & $E_\psi^m$ $\uparrow$     \\ \hline
    \rowcolor{white}
    U²Net                                   & 0.8727                    & 0.8585                    & 0.8210                    & 0.0449                    & 0.8957                    \\ \hline
    \rowcolor{gray!25}
    \textbf{U²Net+ISESS}                     & \textcolor{blue}{0.8869}  & \textcolor{blue}{0.8769}  & \textcolor{blue}{0.8548}  & \textcolor{blue}{0.0404}  & \textcolor{blue}{0.9165}  \\ \hline
    \rowcolor{white}
    BASNET                                  & 0.8702                    & 0.8578                    & 0.8217                    & 0.0476                    & 0.8972                    \\ \hline
    \rowcolor{gray!25}
    \textbf{BASNET+ISESS}                    & \textcolor{blue}{0.8784}  & \textcolor{blue}{0.8630}  & \textcolor{blue}{0.8397}  & \textcolor{blue}{0.0436}  & \textcolor{blue}{0.9057}  \\ \hline
    \rowcolor{white}
    Auto-MSFNet                                  & 0.8662                    & 0.8515                    & 0.8256                    & 0.0425                    & 0.9083                     \\ \hline
    \rowcolor{gray!25}
    \textbf{Auto-MSFNet+ISESS}                    & \textcolor{blue}{0.8703}  & \textcolor{blue}{0.8544}  & \textcolor{blue}{0.8332}  & \textcolor{blue}{0.0417}  & \textcolor{blue}{0.9095}                    \\ \hline\hline
    
    \rowcolor{white}
    \textcolor{red}{\textbf{HKU-IS}}        & $S_m$ $\uparrow$          & $max F_\beta$ $\uparrow$  & $F_\beta^w$ $\uparrow$    & $MAE$ $\downarrow$        & $E_\psi^m$ $\uparrow$     \\ \hline
    U²Net                                   & \textcolor{blue}{0.9183}  & 0.9291                    & 0.8950                    & 0.0311                    & 0.9453                    \\ \hline
    \textbf{U²Net+ISESS}                     & 0.9174                    & \textcolor{blue}{0.9292}  & \textcolor{blue}{0.9085}  & \textcolor{blue}{0.0274}  & \textcolor{blue}{0.9528}  \\ \hline  
    BASNET                                  & \textcolor{blue}{0.9123}  & \textcolor{blue}{0.9264}  & 0.8977                    & 0.0308                    & 0.9476                    \\ \hline
    \textbf{BASNET+ISESS}                    & 0.9110                    & 0.9262                    & \textcolor{blue}{0.9075}  & \textcolor{blue}{0.0278}  & \textcolor{blue}{0.9536}  \\ \hline
    Auto-MSFNet                                  & 0.9148                    & \textcolor{blue}{0.9297}  & 0.9144                    & \textcolor{blue}{0.0255}  & \textcolor{blue}{0.9617}  \\ \hline
    \textbf{Auto-MSFNet+ISESS}                    & \textcolor{blue}{0.9156}  & 0.9270                    & \textcolor{blue}{0.9146}  & \textcolor{blue}{0.0255}  & 0.9603                    \\ \hline
    \end{tabular}
    }
    \caption{Quantitative comparison between \textbf{enhanced} and non-enhanced maps in terms of $S_m$ $\uparrow$, $max F_\beta$ $\uparrow$, $F_\beta^w$ $\uparrow$, $MAE$ $\downarrow$, and $E_\psi^m$ $\uparrow$. \textcolor{blue}{BLUE} indicates the best result between the enhaced and non-enhanced maps.}
    \label{tab:quantitative-results}
\end{table}

\subsection{Qualitative comparisons with non-enhanced maps}\label{subsec:qualitative-comparison}
\begin{figure*}[t!]
    \centering
    \begin{tabular}{c c c c c c c c}
             \includegraphics[width=0.1\textwidth]{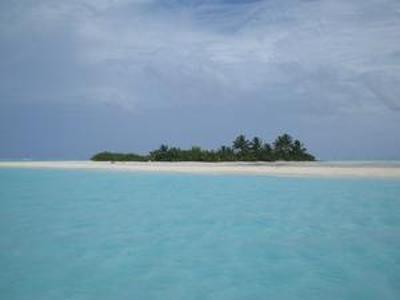} &
             \includegraphics[width=0.1\textwidth]{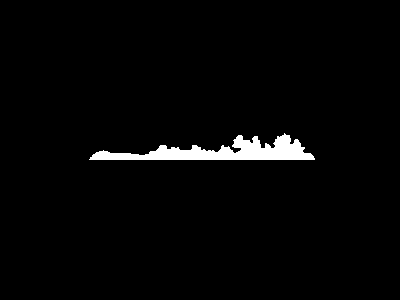} &
             \includegraphics[width=0.1\textwidth]{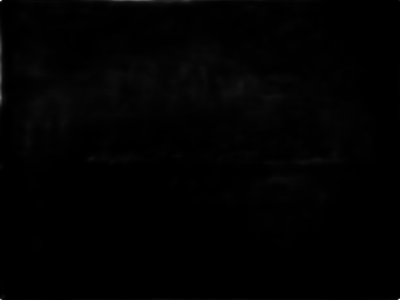} &
             \includegraphics[width=0.1\textwidth]{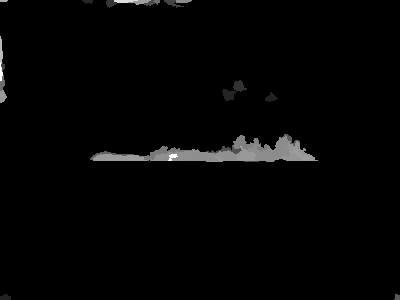} &
             \includegraphics[width=0.1\textwidth]{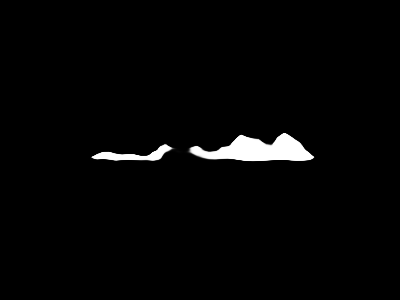} &
             \includegraphics[width=0.1\textwidth]{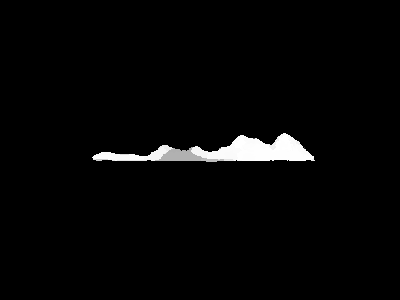} &
             \includegraphics[width=0.1\textwidth]{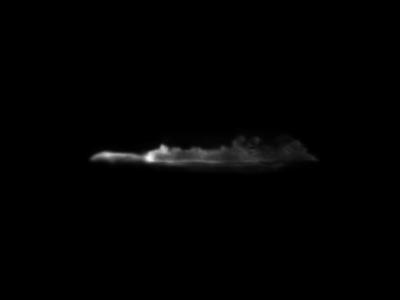} &
             \includegraphics[width=0.1\textwidth]{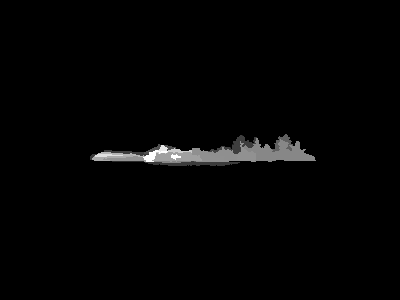} \\
             \includegraphics[width=0.1\textwidth]{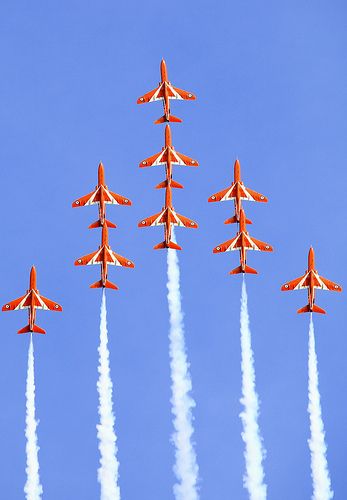} &
             \includegraphics[width=0.1\textwidth]{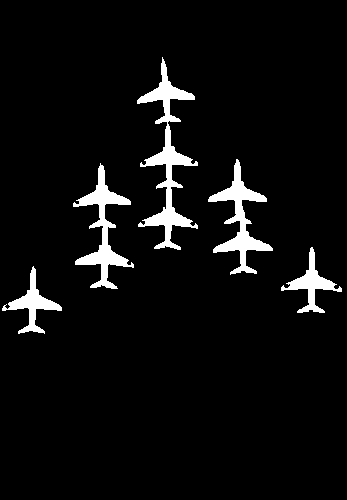} &
             \includegraphics[width=0.1\textwidth]{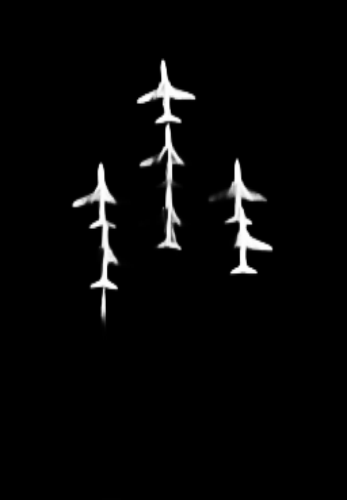} &
             \includegraphics[width=0.1\textwidth]{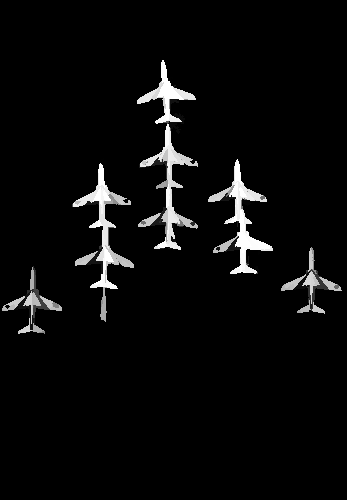} &
             \includegraphics[width=0.1\textwidth]{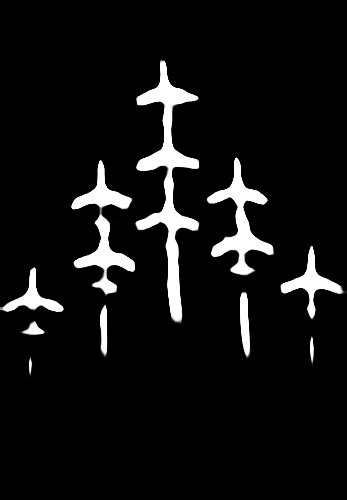} &
             \includegraphics[width=0.1\textwidth]{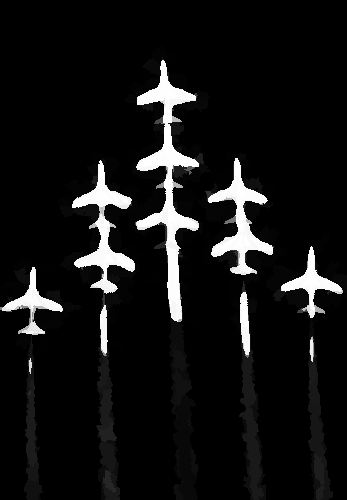} &
             \includegraphics[width=0.1\textwidth]{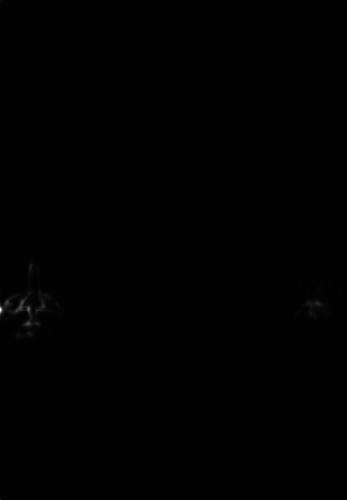} &
             \includegraphics[width=0.1\textwidth]{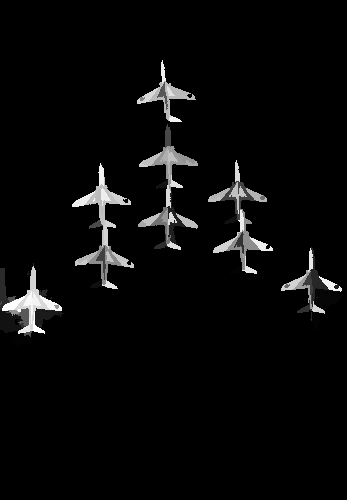} \\
             \includegraphics[width=0.1\textwidth]{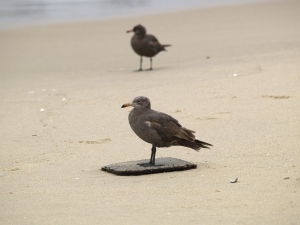} &
             \includegraphics[width=0.1\textwidth]{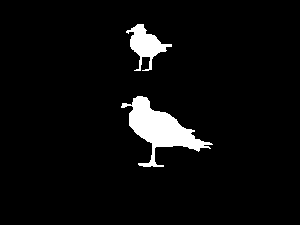} &
             \includegraphics[width=0.1\textwidth]{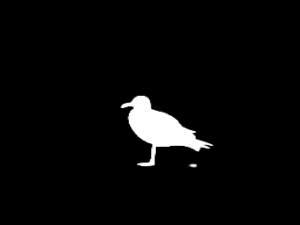} &
             \includegraphics[width=0.1\textwidth]{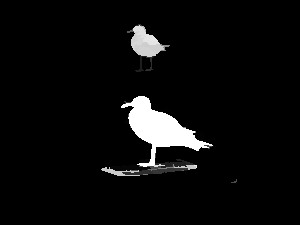} &
             \includegraphics[width=0.1\textwidth]{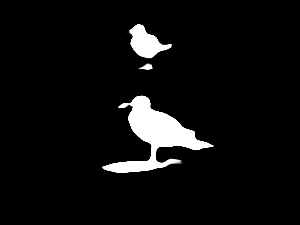} &
             \includegraphics[width=0.1\textwidth]{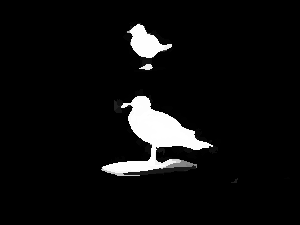} &
             \includegraphics[width=0.1\textwidth]{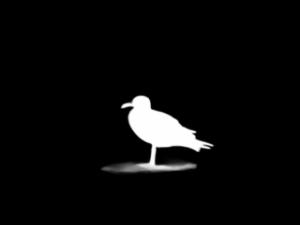} &
             \includegraphics[width=0.1\textwidth]{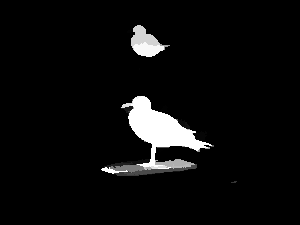} \\
             \includegraphics[width=0.1\textwidth]{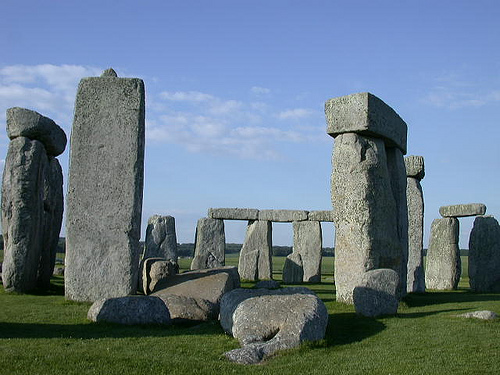} &
             \includegraphics[width=0.1\textwidth]{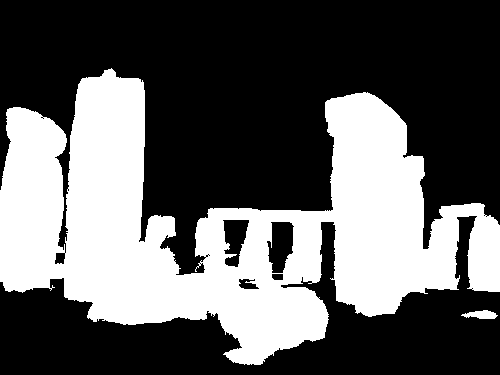} &
             \includegraphics[width=0.1\textwidth]{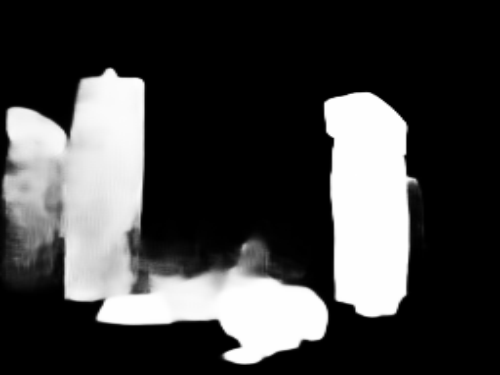} &
             \includegraphics[width=0.1\textwidth]{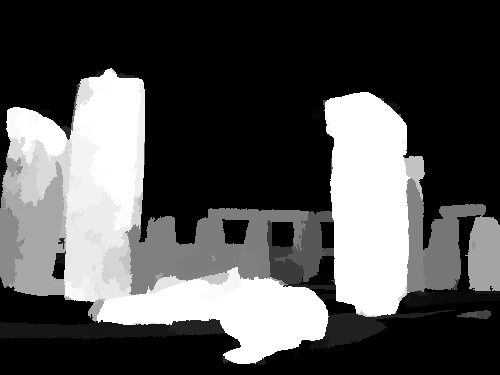} &
             \includegraphics[width=0.1\textwidth]{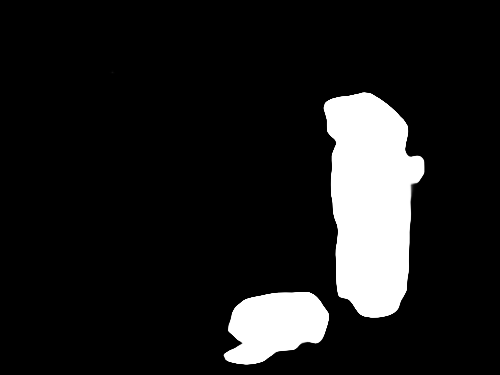} &
             \includegraphics[width=0.1\textwidth]{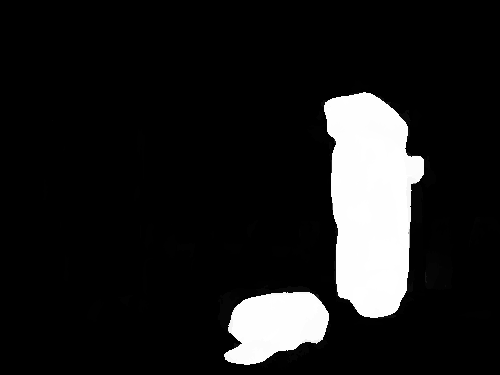} &
             \includegraphics[width=0.1\textwidth]{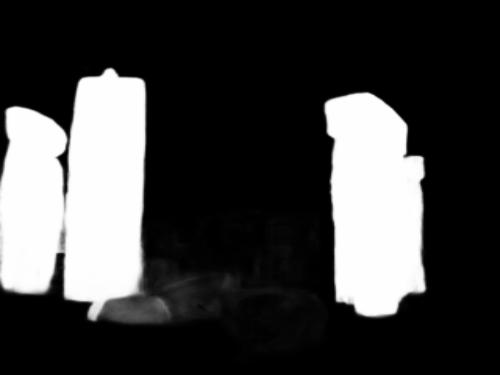} &
             \includegraphics[width=0.1\textwidth]{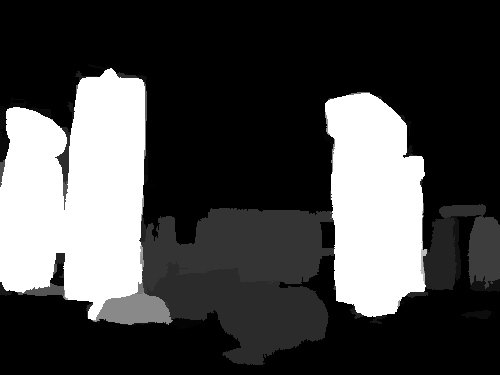} \\
             \includegraphics[width=0.1\textwidth]{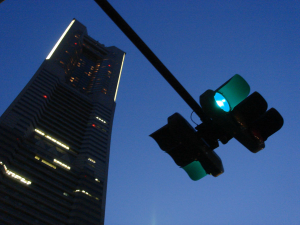} &
             \includegraphics[width=0.1\textwidth]{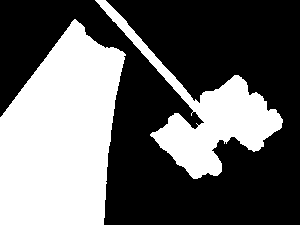} &
             \includegraphics[width=0.1\textwidth]{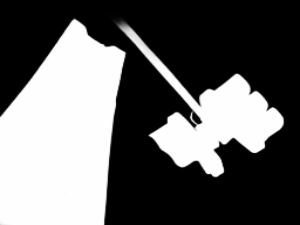} &
             \includegraphics[width=0.1\textwidth]{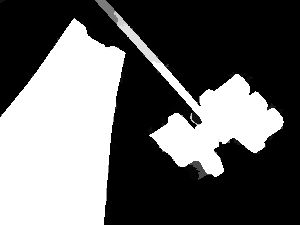} &
             \includegraphics[width=0.1\textwidth]{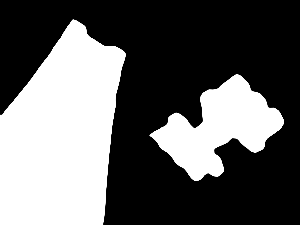} &
             \includegraphics[width=0.1\textwidth]{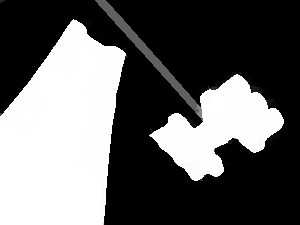} &
             \includegraphics[width=0.1\textwidth]{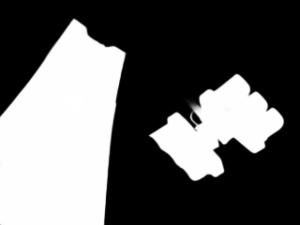} &
             \includegraphics[width=0.1\textwidth]{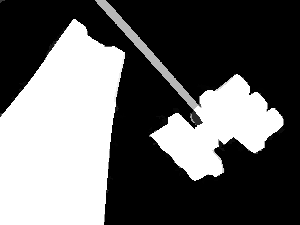} \\
             \includegraphics[width=0.1\textwidth]{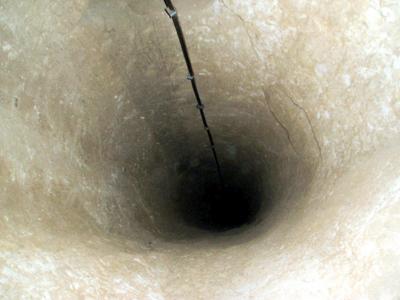} &
             \includegraphics[width=0.1\textwidth]{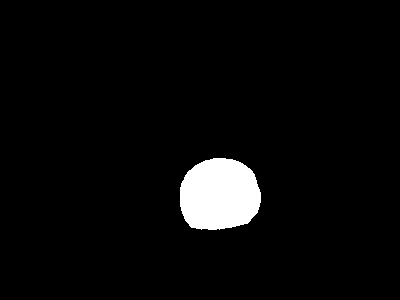} &
             \includegraphics[width=0.1\textwidth]{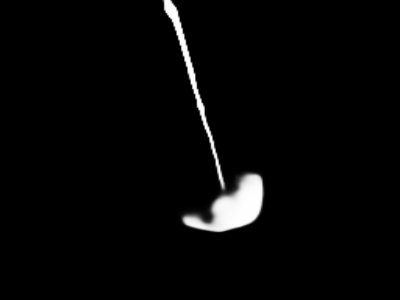} &
             \includegraphics[width=0.1\textwidth]{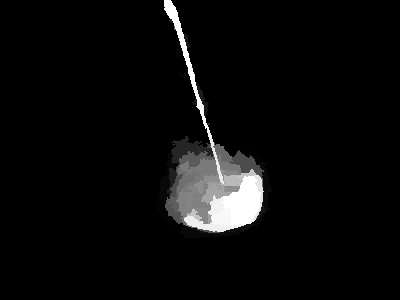} &
             \includegraphics[width=0.1\textwidth]{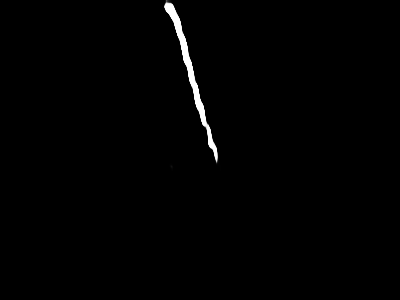} &
             \includegraphics[width=0.1\textwidth]{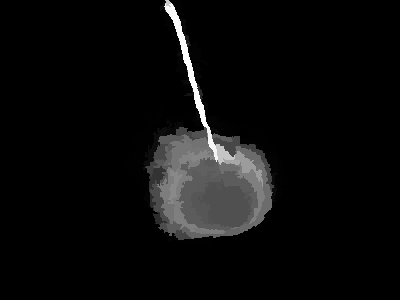} &
             \includegraphics[width=0.1\textwidth]{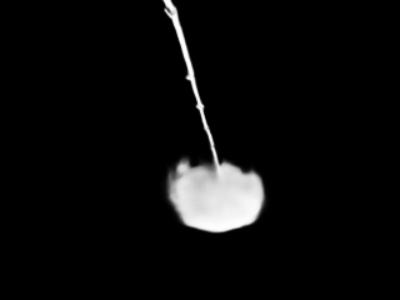} &
             \includegraphics[width=0.1\textwidth]{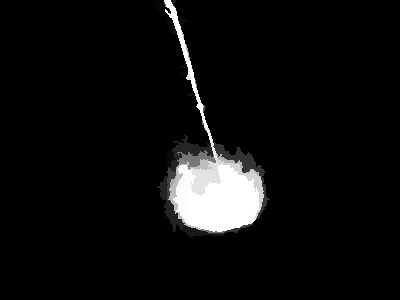} \\
              (a) & (b) & (c) & (d) & (e) & (f) & (g) & (h)
        \end{tabular}
    \caption{Mural of images with their initial saliency maps and their ISESS enhanced version. (a) original image; (b) ground-truth; (c) BASNET \citep{qin2019basnet}; (d) BASNET+ISESS; (e) Auto-MSFNet \citep{miao2021automsf}; (f) Auto-MSFNet+ISESS; (g) U²Net \citep{qin2020u2}; (h) U²Net + ISESS }
    \label{fig:qualitative-mural}
\end{figure*}

To visually evaluate the benefits of enhancing a saliency map using ISESS, we compiled images where ISESS improved different aspects of object representation (Figure \ref{fig:qualitative-mural}).
    In the first two rows, there are images with minimal salient values in a small part of the desired objects (columns (c) and (g), on the first and second row, respectively). Even though the initial map provides little trust (\textit{i.e.}, small saliency values), ISESS created similar resulting maps using all deep models.
    
    In the third and fourth rows, we present examples where ISESS captured most intended objects by extending the saliency value to regions with high similarity to the provided saliency. Note, however, that ISESS was unable to improve (e), even though (e) is a high precision map. Because the map is almost binary and most salient regions are taken as background queries, ISESS could not highlight the other salient regions.
    
    The bottom four rows present images where ISESS was able to complete partially salient objects. Excluding the last row, each row shows simpler objects that the deep models failed to capture fully. In particular, in (e) and (g), ISESS could properly include a narrow stem coming out of the traffic light, even though narrow structures can be complex during superpixel delineation. The last image presents a successful case on a difficult challenge due to the radiating shadow coming out of the hole (which makes it hard to delineate adequate borders), especially on (e) where the initial map gave no part of the desired object.

\section{Conclusion}\label{sec:conclusion}
We presented a hybrid model for saliency enhancement that exploits a loop between  superpixel segmentation and saliency enhancement for the first time. ISESS delineates superpixels based on object information, as represented by an input saliency map, and improves the input saliency map by computing feature similarity between superpixels and queries. It exploits multiple scales of superpixel representation and integrates the intermediate saliency maps by cellular automata. Experimental results on five public SOD datasets demonstrate that ISESS can consistently improve object representation, especially in the following cases: (a) partially salient objects and (b) images with multiple salient objects with only a few captured by the deep model. Although better superpixel descriptors can be used for superpixel similarity, we focused on a simple color descriptor to illustrate how well deep models can be assisted by image-intrinsic information to improve their results.

We intend to explore ISESS enhanced maps for interactive- and co-segmentation tasks. The goal will be to assist humans when annotating images with fewer interactions, using the highly-trusted human-provided object location to define the precise background and foreground queries, combined with robust deep models to capture finer features. Another characteristic inherent in ISESS is the superpixel improvement over time. Although this aspect was not in focus in this paper, future work includes further evaluation of the enhancement loop's impact on superpixel segmentation and how the final superpixel map can assist interactive object segmentation.

\section*{Acknowledgement}
This work was supported by ImmunoCamp, CAPES, CNPq (303808/2018-7) and \href{http://data.crossref.org/fundingdata/funder/10.13039/501100001807}{FAPESP} (2014/12236-1).

%\FloatBarrier

\bibliographystyle{model2-names}\biboptions{authoryear}
\bibliography{main}

%\vspace{2\baselineskip}
%\par\noindent {\bf Alexandre Xavier Falcao}\
%is a professor at the Institute of Computing (IC), University of Campinas (UNICAMP). He has worked on machine learning, computer vision, image processing, and data visualization at the IC-UNICAMP with research productivity fellowships from the Brazilian National Council for Scientific and Technological Development (CNPq) since 1998.
%\vspace{2\baselineskip}

%\par\noindent {\bf Leonardo de Melo Joao}\
%is currently pursuing a Ph.D. in Computer Science at University of Campinas (UNICAMP), SP, Brazil. He received a M.Sc degree in Computer Science from the University of Campinas (UNICAMP) in 2020. His research interests include machine learning and image processing, with an emphasis on saliency estimation and image segmentation.

\end{document}